\documentclass[final]{cvpr}

\usepackage{times}
\usepackage{epsfig}
\usepackage{graphicx}
\usepackage{amsmath}
\usepackage{amssymb}

\usepackage{bbm}
\usepackage{xcolor}
\usepackage{float}
\usepackage{booktabs}       %
\usepackage{multicol}
\usepackage{multirow}
\usepackage{caption}
\usepackage{subcaption}
\usepackage{dsfont}
\usepackage{flushend}

\newcommand{\myparagraph}[1]{\vspace{2mm}\noindent\textbf{#1}}

\usepackage[pagebackref=true,breaklinks=true,colorlinks,bookmarks=false]{hyperref}

\def\cvprPaperID{10382} %
\def\confYear{CVPR 2021}

\begin{document}

\title{Ensembling with Deep Generative Views}

\author{
Lucy Chai\textsuperscript{1,2}\hspace{4mm} Jun-Yan Zhu\textsuperscript{2,3}\hspace{4mm} Eli Shechtman\textsuperscript{2}\hspace{4mm} Phillip Isola\textsuperscript{1}\hspace{4mm} Richard Zhang\textsuperscript{2} \vspace{1mm}\\
\textsuperscript{1}MIT\hspace{20mm}\textsuperscript{2}Adobe Research\hspace{20mm}\textsuperscript{3}CMU \vspace{-.5mm}\\
{\tt\small \{lrchai, phillipi\}@mit.edu \hspace{0.5pt} \{elishe, rizhang\}@adobe.com \hspace{0.5pt}  \hspace{0.5pt} junyanz@cs.cmu.edu  }
}

\maketitle

\begin{abstract}

Recent generative models can synthesize ``views'' of artificial images that mimic real-world variations, such as changes in color or pose, simply by learning from unlabeled image collections. Here, we investigate whether such views can be applied to real images to benefit downstream analysis tasks such as image classification. Using a pretrained generator, we first find the latent code corresponding to a given real input image. Applying perturbations to the code creates natural variations of the image, which can then be ensembled together at test-time. We use StyleGAN2 as the source of generative augmentations and investigate this setup on classification tasks involving facial attributes, cat faces, and cars. Critically, we find that several design decisions are required towards making this process work;
the perturbation procedure, weighting between the augmentations and original image, and training the classifier on synthesized images can all impact the result. Currently, we find that while test-time ensembling with GAN-based augmentations can offer some small improvements, the remaining bottlenecks are the efficiency and accuracy of the GAN reconstructions, coupled with classifier sensitivities to artifacts in GAN-generated images.

\end{abstract}

\vspace{-5mm}
\section{Introduction}

Image datasets are the backbone of learning-based vision problems, but images are only sparsely-sampled discrete snapshots of the underlying continuous world. %
However, recent generative adversarial networks (GANs)~\cite{goodfellow2014explaining} have shown promise in learning to imitate the real-image manifold, mapping random samples from a latent distribution to realistic image outputs. A heavily exploited property of these models is that the latent space is locally smooth: samples nearby in latent space will appear perceptually similar in image space~\cite{radford2015unsupervised}. Therefore, GANs can be viewed as a type of ``interpolating mechanism'' that can blend and recombine images in a continuous manner. From individual image samples, can we use a GAN to generate nearby alternatives on the image manifold, or ``views,'' giving us unlimited variants of a given image?

\begin{figure}[t!]
  \centering
  \includegraphics[width=0.48\textwidth]{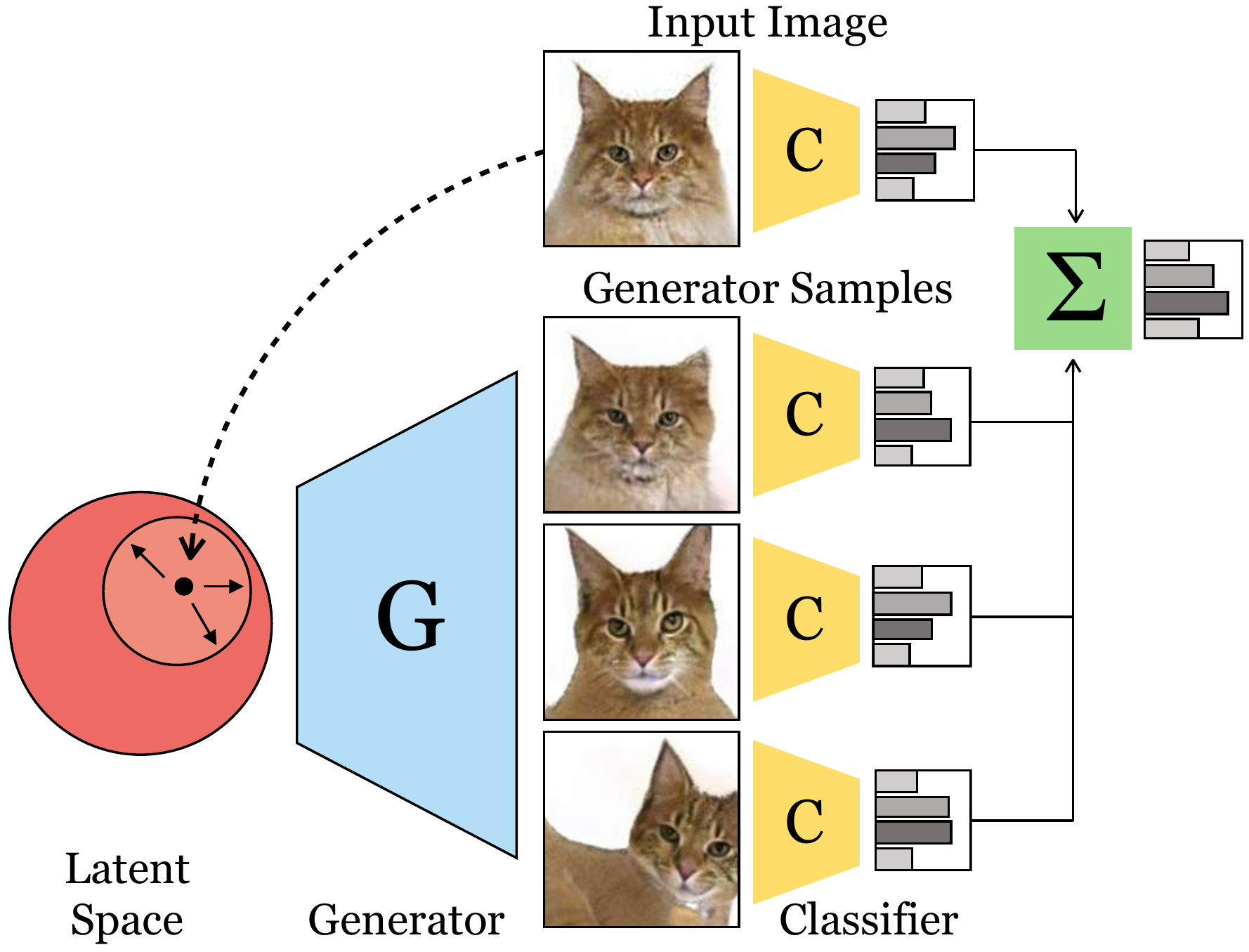}
  \vspace{-1em}
  \caption{\small We project an input image into the latent space of a pre-trained GAN and perturb it slightly to obtain modifications of the input image. These alternative views from the GAN are ensembled at test-time, together with the original image, in a downstream classification task.
  \label{fig:teaser}}
  \vspace{-0.2in}
\end{figure}

Here, we investigate using GAN outputs as test-time augmentation for classification tasks. In the standard classification pipeline, passing an image through a trained classifier yields predictions of the image belonging to one of several classes. However, performance is often improved using more than one sample -- if we have multiple views of an image, we can use the classifier to obtain predictions for each view and average the results together as an ensemble. The classic approach to generating additional views has been to crop the image at different locations before classification. Using a GAN, we have an orthogonal, data-driven method of generating additional views of a given image, such as altering its pose, shape, or color based on the directions of variation that a GAN learns. %

A secondary advantage of unconditional GANs is that they can be trained on image collections without requiring image labels. As data labeling is often vastly more expensive than data collection, GANs learn from much larger datasets compared to tasks involving manual annotation, such as classification. Training on large datasets allows several interesting properties to emerge, where the generator learns meaningful variations in data 
without requiring an explicit training objective to do so~\cite{harkonen2020ganspace,shen2020closed,collins2020editing}. 

A challenge in using GANs to generate augmented samples is the potential domain gap between real images and GAN outputs -- the generated samples must be of sufficient quality to be used in classification tasks, and must adequately reconstruct the target image sample while preserving the relevant visual patterns for accurate discrimination. If this condition is not met, the classifier may behave differently on the generated samples than the natural images, which is undesirable for data augmentation. To reconstruct image samples using the GAN, we use a hybrid encoder and optimization approach~\cite{zhu2016generative}: the encoder network initializes the latent code, which is further optimized to improve the similarity between the target image and the reconstruction. In addition to high-quality reconstruction, we also desire that the generated variations of an image do not cross classifier boundaries; \ie, it cannot modify visual appearance that affect its classification. To this end, we experiment with a variety of possible image modifications using the GAN generator, ranging from coarse pose and shape changes to finer-grained color changes. 

Using the recent StyleGAN2 generator~\cite{karras2019analyzing}, we apply our method on several classification tasks involving facial attributes~\cite{liu2018large,karras2017progressive}, cat faces~\cite{parkhi2012cats}, and cars~\cite{krause2013object}. Given the relative simplicity of the face domain, we find test-time ensembling with GAN samples helps even when the classifier is trained only on real images; however, training the classifier on generated samples offers further improvements, particularly for the more difficult car and cat domains. 
Code is available on our website: \url{https://chail.github.io/gan-ensembling/}.

\section{Related Work}

\noindent \textbf{Latent manipulation in GANs.}
Advances in generative adversarial networks have allowed them to create increasingly realistic images~\cite{denton2015deep,radford2015unsupervised,karras2017progressive,brock2018large,zhang2019self,karras2019style,karras2019analyzing}, and furthermore, the generated outputs mimic the variations found in their training data. 
For example, Radford et al.~\cite{radford2015unsupervised} demonstrate %
linear separability in latent space and use it to modify attributes of interest in GAN-generated samples.
This linear separability has been exploited in a number of subsequent works, including for faces~\cite{shen2019interpreting}, for camera attributes~\cite{jahanian2019steerability,plumerault2020controlling}, and for quantifying disentanglement of the latent space~\cite{karras2019style}. While these manipulations require a supervised or self-supervised objective, another direction of interest involves learning meaningful editing directions without direct supervision. One set of approaches aims to uncover primary directions of variation in an intermediate latent space~\cite{harkonen2020ganspace,shen2020closed,collins2020editing}, %
while another enforces distinctness of optimized directions during training~\cite{voynov2020unsupervised}. Apart from linear edit directions, recent works~\cite{zhang2020image, pan2020gan2shape} learn latent editing operations that control the 3D appearance of generated images.
Interestingly, the architecture of a GAN itself may lend itself to natural edits; for example, StyleGAN, which we use~\cite{karras2019style,karras2019analyzing}, provides a hierarchical latent code that controls visual patterns at various scales.

\vspace{2mm} \noindent \textbf{GANs for real image editing.}
While GANs are able to generate interesting variations of synthetically generated images, one is often more interested in editing a given \textit{real} image. To do this, we first need to find the latent code that best reconstructs the image~\cite{zhu2016generative,brock2016neural}, %
a challenging problem due to the generator's inherent limitations and the complex optimization landscape through a deep network.
Optimizing in an intermediate layer or an expanded latent space allows the generator to reconstruct a greater variety of real images~\cite{abdal2019image2stylegan,abdal2020image2stylegan++,bau2018gan}, which relaxes the former restriction. However, one runs the risk of overfitting or drifting off the manifold of natural latent codes;
as such, works~\cite{zhu2020domain,wulff2020improving} propose regularizers to constrain the optimized code to the latent manifold. To speed up the optimization process, several methods~\cite{zhu2016generative,zhu2020domain,bau2019seeing,richardson2020encoding} propose to train an encoder to initialize latent optimization. To further improve the generator's flexibility and reconstruction capability, Pan et al.~\cite{pan2020exploiting} and Bau et al.~\cite{bau2020semantic} finetune the generator weights towards a particular target image, while Huh et al.~\cite{huh2020transforming} propose to spatially transform the image due to the generator's inherent biases.
As a fast and accurate image projection is critical to our pipeline, we directly draw upon the insights from these works in our method.

\vspace{2mm} \noindent \textbf{Data augmentation with generative models.}
Generative models' ability to create realistic variations of images opens up the possibility of using them as data augmentation in downstream tasks, as acquiring images from a generative model is vastly cheaper than collecting additional data. Several works investigate using generative samples as data augmentation in an image translation setting, in which an autoencoder-style network is trained to produce variations of an input image, on a variety of domains such as face attribute editing~\cite{viazovetskyi2020stylegan2}, gaze estimation~\cite{shrivastava2017learning}, emotion classification~\cite{zhu2018emotion}, and honeybee tracking~\cite{sixt2018rendergan}. Rather than taking an image-to-image translation approach, we use \textit{projection into a pretrained generator}, which allows us to use emergent variations from this generator rather than directly training for a specified transformation. Instead of a generative model for images, Ratner et al.~\cite{ratner2017learning} formulates a generator to select from a sequence of predefined image augmentations used to train a downstream classifier. Specifically in the medical community, synthesizing additional images to augment limited-size datasets has been beneficial for classification~\cite{perez2018data,wu2018conditional,frid2018synthetic,hou2017unsupervised,salehinejad2018generalization} or segmentation tasks~\cite{liu2019pixel,chaitanya2020semi,guibas2017synthetic}. However, using GANs for data augmentation is limited by the possible domain gap between generated samples and dataset images~\cite{ravuri2019seeing,ravuri2019classification}. Studies show that despite their rapid improvement, the generative models do, in fact, still exhibit artifacts, exploitable for reliable detection of GAN-generated imagery~\cite{wang2020cnn,chai2020makes,yu2019attributing,zhang2019detecting,frank2020leveraging}. Outputs from a generative model can also be used in adversarial settings: Samangouei et al.~\cite{samangouei2018defense} demonstrate projecting adversarial images through a generator on MNIST and Fashion-MNIST, while other works~\cite{shen2017ape,jalal2017robust,gowal2020achieving} train models using the generator to improve robustness. Augmentation using GANs can be applied at training time for robustness or test-time for ensembling;  while we primarily focus on the latter, concurrent works ~\cite{mao2020generative,sauer2021counterfactual} investigate the benefits of the former approach while~\cite{tritrong2021repurposing} investigates intermediate GAN representations for few-shot segmentation.

\section{Method}

Our goal is to leverage a generative model to synthesize useful variations of a given image. As summarized in Fig.~\ref{fig:teaser}, the first step is to ``project'' the image to the latent code of a generator. From there, we explore different methods for producing variations. Finally, we show how these variations can be effectively used by a downstream classifier.

\subsection{GAN preliminaries}

A generative network maps a low-dimensional code $z$
to an image $x$.
Specifically, we use the StyleGAN2 generator~\cite{karras2019analyzing}. A useful property of the architecture is the intermediate feature space $w\in\mathds{R}^{B\times D}$, which contains $B$ blocks of dimension $D$, designed to control the ``style'' of the image~\cite{karras2019analyzing}.\footnote{Previous work~\cite{abdal2019image2stylegan} refers to this as the $w^{+}$ space, with a separate, more constrained $w$ space. We do not need to make this distinction, so we refer to it as $w$ for simplicity.} Earlier and later blocks affect the image on coarser and finer scales, respectively. Combining the mapping network $M$ and backbone generator $G$ produces image $x=G(w)=G(M(z))$. Previous work~\cite{abdal2019image2stylegan} finds that the intermediate $w$ space is better able to represent images than the original code $z$, while moving in this space offers controllable and interesting effects~\cite{harkonen2020ganspace}. As such, work on StyleGAN2 inversion~\cite{abdal2020image2stylegan++,shen2019interpreting,zhu2020domain} typically uses this intermediate space, and we refer to it as the ``latent code''. Following StyleGAN~\cite{karras2019style}, we subdivide $w\in\mathds{R}^{B\times D}$ into ``coarse'' styles (first four style codes), ``middle'' styles (next six style codes), and ``fine'' styles  (the remaining style codes). As we find that changing the ``middle'' layers can alter object identity, we focus on modifying the ``coarse'' and ``fine'' styles to produce image variations.

\subsection{Projecting Images into GAN Latent Space}

To edit a real image $x$, we must first find the latent code $w$ that generates the image.
As an exact match cannot usually be found, this problem is relaxed to finding the closest image
by solving the following optimization problem over an image distance metric:
\begin{equation}
\begin{split}
w^* &= \arg\min_w \mathcal{L}_\text{img}(G(w), x) \\
 &= \arg\min_w \, \, \ell_1(G(w),x) + \mathcal{L}_\text{percep}(G(w), x).
\end{split}
\end{equation}
\noindent We use $\ell_1$ and LPIPS~\cite{zhang2018unreasonable} distances, denoted as $\mathcal{L}_\text{percep}$. Solving this projection problem directly via optimization is challenging due to the difficult optimization landscape, and tends to heavily depend on the initialization. This has been the subject of active research~\cite{zhu2016generative,abdal2019image2stylegan,bau2019seeing,abdal2020image2stylegan++,huh2020transforming}. Furthermore, optimization can be slow to converge to a reasonable reconstruction of the target image $x$. We follow best practices, with the needs of our problem in mind. Specifically, a projection algorithm that is very computationally-intensive is not tenable for our use-case, as our approach must be feasible across a full dataset.
To balance reconstruction quality and optimization time, we use a two-step approach to project images into the GAN latent space.

\begin{figure*}[t!]
  \centering
    \includegraphics[width=\textwidth]{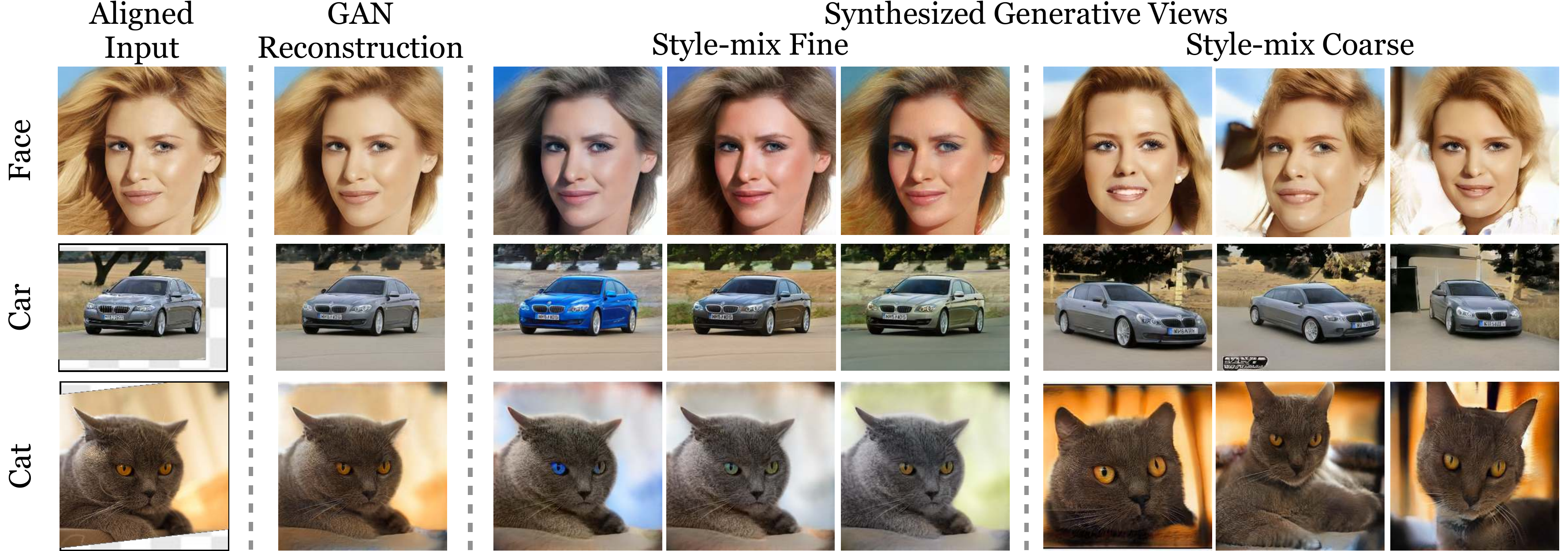}
  \vspace{-2em}

    \caption{\small Synthesizing deep generative views. We first align (Aligned Input) and reconstruct an image by finding the corresponding latent code in StyleGAN2~\cite{karras2019style} (GAN Reconstruction).
    We then investigate different approaches to produce image variations using the GAN, such as style-mixing on fine layers (Style-mix Fine), which predominantly changes color, or coarse layers (Style-mix Coarse), which changes pose. We show additional perturbations
    in supplementary material.
  \label{fig:qualitative_perturbation}}
  \vspace{-0.2in}
\end{figure*}

\vspace{1mm} \noindent \textbf{Preprocessing by alignment.} As GANs have a tendency to accentuate spatial biases and generate centered objects~\cite{huh2020transforming}, we shift the image to center the target object prior to image projection (Fig.~\ref{fig:qualitative_perturbation} left).
As this may cause some missing pixels around the edges of the image, we use spatial masks in the projection objective, so that unknown edge pixels do not contribute to the loss. Please refer to the supplement for additional details.

\vspace{1mm} \noindent \textbf{Encoder initialization.} As a first step, we train an encoder network $E$, to initialize the latent code with a single forward pass. The encoder network is trained using the objective:
\begin{equation}
\mathcal{L}(x, w, E) = \mathcal{L}_\text{img}(x, G(E(x))) + \lambda \mathcal{L}_\text{latent}(w, E(x)).
\end{equation}

\noindent The $\mathcal{L}_\text{latent}$ is an $\ell_2$ reconstruction loss for supervision. For this stage, we set $\lambda=1.0$. The encoder $E$ is trained across randomly drawn, corresponding latent codes and images:

\begin{equation}
E^{*} = \arg\min_E \mathds{E}_{w,x} \mathcal{L}(x, w, E).
\end{equation}

\vspace{1mm} \noindent \textbf{Iterative optimization.} 
Given a specific image $x$, a forward pass through the trained encoder yields an initialization. A closer match can be obtained by further optimization:

\begin{equation}
w^{*} = \arg\min_w \mathcal{L}(x, w, E^{*}).
\end{equation}

\noindent Note that nearly the same objective can be used as for training the encoder. We reduce $\lambda$ to 0.5, as here, the latent loss serves as a regularizer to the encoder-provided initialization, rather than as ground truth supervision. We optimize for 500 iterations using an L-BFGS optimizer~\cite{liu1989limited} taking 30-45 seconds per image on a V100 GPU, depending on the generator resolution. As this optimization is performed for each image, we limit the optimization time to be short enough so that computation is tractable over the dataset, yet long enough to reasonably match the target image.

\subsection{Image Augmentations using Pretrained GANs}
\label{sec:method_perturbation}
Once we optimize for a latent code $w^{*}$ matching a given target image, we then perturb it to obtain image variations. We try local perturbations using an isotropic Gaussian and PCA directions, along with a ``style'' swapping operation.

\vspace{1mm} \noindent \textbf{Isotropic Gaussian.} One approach is to simply sample from an isotropic Gaussian ball centered at the optimized latent:
\begin{equation}\label{eqn:perturb_isotropic}
    \tilde{w} \sim \mathcal N(w^{*}, \sigma I),
\end{equation}
where $\sigma$ scales variance of the sampled points. We obtain hyperparameter $\sigma$ by cross-validating over the validation set, and ultimately report on the test set. Note that using $\sigma=0$ yields the reconstructed image, while a large $\sigma$ corresponds to randomly drawing a generated image with little regard to the original image $x$. We conduct experiments adding noise to either the ``coarse'' or ``fine'' style codes.

\vspace{1mm} \noindent \textbf{Principal directions.} Alternatively, we also experiment with sampling according to principal directions; these directions were found to correspond well with interpretable controls in GANspace~\cite{harkonen2020ganspace}. We perform PCA on random samples in the latent space to obtain principal directions $\hat{v}_d$, which are unit vectors, and eigenvalues $\lambda_d$. To produce perturbations, we randomly sample a principal component direction $d$ uniformly among $n$ principal components, $d \sim \mathcal{U}[1, n]$, and a perturbation factor $\beta$ from a uniform distribution $\beta \sim \mathcal{U}[-\sigma, \sigma]$. In practice, as only the top directions have a visible effect, we restrict ourselves to the top $n=20$ principal components:

\begin{equation}\label{eqn:perturb_pca}
    \tilde{w} = w^{*} + \beta \lambda_d \hat{v}_d. %
\end{equation}
We modify either the ``coarse'' or ``fine'' style codes while holding the remaining layers fixed.

\vspace{1mm} \noindent \textbf{Style-mixing.}\label{eqn:perturb_stylemix}
Recall that a property of StyleGAN2 is that  $w^{*}\in\mathds{R}^{B\times D}$ where the early layers correspond to the coarse visual patterns, while later layers correspond to fine details. The third method of perturbation corresponds to ``style-mixing'', which swaps in a randomly generated latent code $w$ at some granularity (e.g., the ``fine'' styles), while preserving the remaining layers from the optimized $w^{*}$ (\cite{karras2019analyzing}; Fig.~\ref{fig:qualitative_perturbation}-right).
Geometrically, rather than jittering locally, this corresponds to jumping onto a vertex on a hypercube defined by the two latent codes. Visually, this enables us to achieve greater changes in the appearance of the perturbed output, compared to local jittering.

\subsection{Ensembling deep generative views}

In typical classification, feeding an image $x$ through a trained classifier $C$ yields the prediction logits for that image $y = C(x)\in\mathds{R}^{L}$, where $L$ is the number of classes.

\vspace{1mm} \noindent \textbf{Ensembling.}\label{sec:ensemble_weight} By perturbing the optimized latent codes, we obtain additional samples, which can be ensembled at test-time. We jitter the optimized latents to obtain a series of latent codes $\tilde{w}_1 \cdots \tilde{w}_N$ and then run these codes through the generator to obtain variations of the target image: $G(\tilde{w}_1)\cdots G(\tilde{w}_N)$ to as inputs to a trained classifier $C$. To ensemble the classifier predictions, we find that an appropriate weighting between the original image $x$ and the generated samples $\{G(\tilde{w}_n)\}$ improves results. Therefore, we ensemble the end classifier decision using:
\begin{equation}\label{eqn:ensemble_weight}
    y_\text{ens} = (1-\alpha) C(x) + \frac{\alpha}{N}\sum_{n=1}^N C(G(\tilde{w}_n)).
\end{equation}
The hyperparameter $\alpha \in [0, 1]$ is selected using the validation data, and the optimal value in validation is applied to the test partition. Note that $\alpha=0$ means that the perturbations actually reduce validation performance. This could be due to (a) a poor reconstruction of the input image or (b) overaggressive perturbations that do not preserve the property of interest in the image. On the other hand, $\alpha=1$ means that 
the ensembled variations are entirely able to capture necessary properties for the classification task.

\vspace{1mm} \noindent \textbf{Classifier training.} Due to the domain gap between real images and their generated counterparts, we find that training the classifier on generated images often improves results. Given tuples of images, optimized latents, and labels $(x, w^*, y)$, we train the classifier using images generated from the latents and standard cross entropy loss $\mathcal{L}_{\text{cls}}$:
\begin{equation}\label{eqn:training_classifier}
    C^* = \arg \min_C\mathcal{L}_{\text{cls}}(C(G(w^*), y)).
\end{equation}
We also experiment with training on perturbed $\tilde{w}$ using the methods in Sec.~\ref{sec:method_perturbation}, thus replacing  $C(G(w^*))$ with $C(G(\tilde{w}))$ in Eqn.~\ref{eqn:training_classifier}. We provide details on classifier architecture and training parameters in supplementary material.

\section{Experiments}

We use the previously described approach of projecting images into a GAN's latent space and then modifying the latent code to create alternative views. We conduct experiments on facial attribute, car, and cat face classification.

\subsection{Facial Attribute Classification}

Human faces are one of the most successful domains for recent GAN architectures, as aligned faces reduce the amount of variation that the GAN needs to model. Furthermore, large datasets exist of labeled facial attributes, which can be subsequently used for classification. We start by investigating this setting by pairing the StyleGAN2~\cite{karras2019analyzing} face model with the CelebA-HQ dataset~\cite{liu2018large,karras2017progressive}, containing 30,000 images with 40 labeled binary attributes. 

\begin{figure}[ht!]
  \centering
  \includegraphics[width=\linewidth]{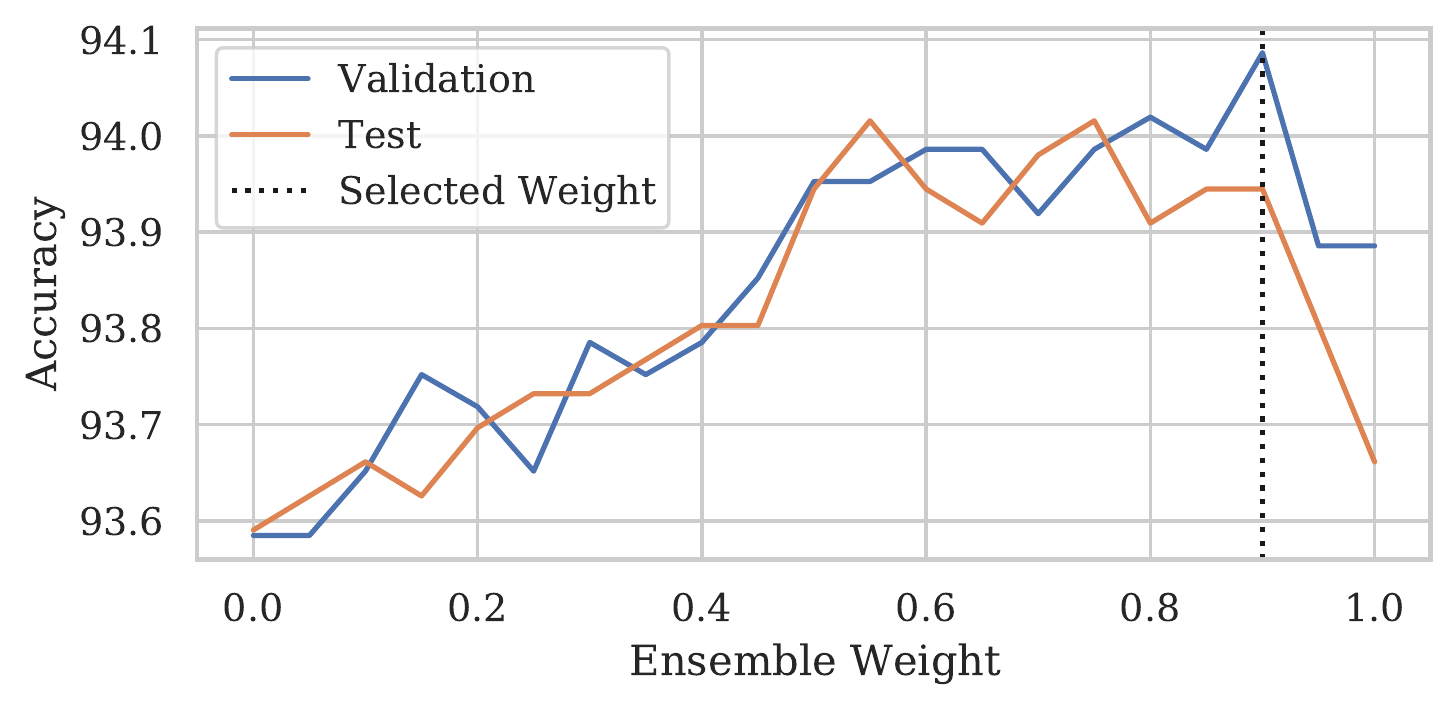}
  \vspace{-8mm}
  \caption{\small Selecting ensemble weight $\alpha$ by cross-validation; $\alpha=0$ corresponds to the standard test accuracy, while $\alpha=1$ corresponds to discarding the original image. We find that often using an intermediate value of $\alpha$ yields the highest accuracy, trading off the additional views provided by the GAN with potential imperfection in reconstruction; we show validation accuracy as a function of $\alpha$ in blue, and test accuracy in orange for the Smiling attribute. We select the value of $\alpha$ based on the validation split and apply the same value to the test split (shown in the dotted black line); note that this may not be the same as the optimal $\alpha$ for the test split. 
  \label{fig:ensemble_weight}}
\end{figure}

\myparagraph{Ensemble weighting.} As described in Sec.~\ref{sec:ensemble_weight}, a soft weighting between the real image and its GAN-generated variants is necessary. Since the GAN reconstructions may not exactly approximate the ground-truth image, this causes the classifier to behave differently on the real images compared to their GAN-generated counterparts. We visualize this effect in Fig.~\ref{fig:ensemble_weight}, where we plot accuracy as a function of the ensemble weight $\alpha$ and find that an intermediate weighting yields the highest accuracy.

\begin{figure}[ht!]
  \centering
  \includegraphics[width=\linewidth]{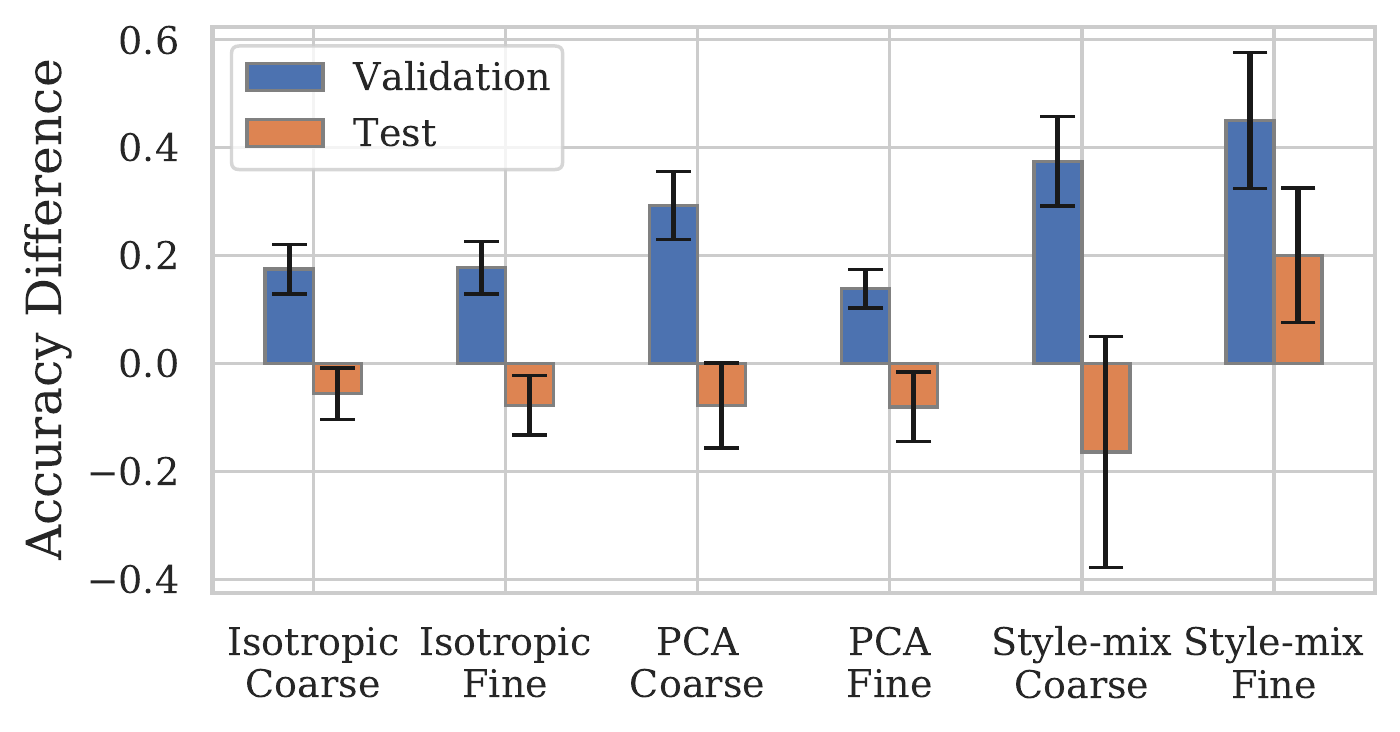}
  \vspace{-8mm}
  \caption{\small Investigating GAN perturbation methods. On a subset of 12 face attributes, we first investigate test-time ensembling using each type of latent code perturbation. The ensemble weight hyperparameter $\alpha$ is selected from the validation split. Therefore, we find that all types of perturbation can increase validation accuracy. However, the style-mixing perturbation on fine layers performs the best at both validation and test time, and only the fine style-mixing operation yields robust improvements at test time. Error bars indicate standard error over the 12 attributes. 
  \label{fig:face_aug_types}}
  \vspace{-4mm}
\end{figure}

\myparagraph{GAN perturbation variations.}
Next, we investigate the effect of the different types of perturbations -- isotropic Gaussian perturbations (Eqn.~\ref{eqn:perturb_isotropic}), principal component directions (Eqn.~\ref{eqn:perturb_pca}), or swapping layers of the optimized latent code (style-mixing; Sec.~\ref{eqn:perturb_stylemix}). We start with a subset of 12 out of 40 facial attributes, and investigate each type of perturbation on the resulting ensemble classification accuracy. In Fig.~\ref{fig:qualitative_perturbation}, we show qualitative examples of the latent code perturbations for the style-mixing operation on fine and coarse layers, and visualize the remaining types of perturbations in the supplementary material. While perturbations in the coarse layers preserve color attributes, they do not always preserve the facial identity. We find that style-mixing in the fine layers offers the most robust improvements at test time; while all types of perturbations offer improvements on the validation split (using the previous ensemble weight $\alpha$ selected on validation data), only the fine style-mixing method offers reliable improvements at test time (Fig.~\ref{fig:face_aug_types}). Therefore, we primarily focus on this type of perturbation for the remaining face experiments.

\begin{figure}[ht!]
  \centering
  \hspace{0.2in}
  \includegraphics[width=\linewidth]{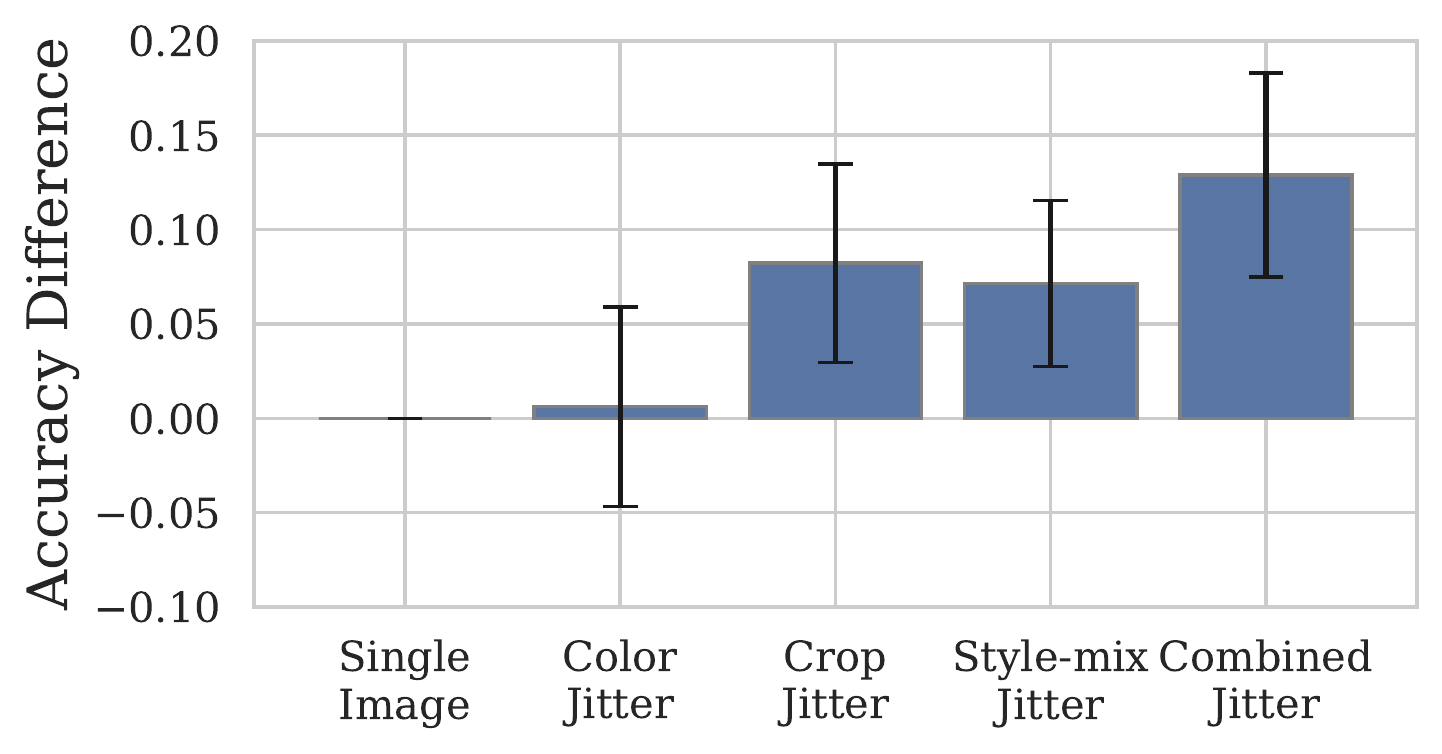}
  \vspace{-8mm}
  \caption{\small Traditional vs. generative ensembling. Averaged over 40 facial attributes, we plot the accuracy increase from test-time ensembling with various forms of augmentation. While ensembling with image augmentations (Color, Crop) or deep generative views (Style-mix) improves over no ensembling (Single Image), more importantly, they provide \textit{orthogonal benefits}; combining them works best. Errorbars show standard error over 40 attributes. 
  \label{fig:face_testaug}}
  \vspace{-10mm}
\end{figure}

\myparagraph{Comparing to image augmentations.} Standard image augmentations used during training, such as random flipping or cropping, can also be ensembled together at test time to obtain a more robust classification result. We compare the GAN-based augmentations to these image-based augmentation techniques, including (1) color jitter with horizontal flip or (2) spatial jitter with horizontal flip (as faces are aligned, we use small spatial perturbation by resizing to 1056px before taking a random 1024px crop). Averaged across all 40 facial attributes, we find that ensembling with color jitter offers small improvements over single-image classification. Adding the style-mixing perturbation provides similar benefits to spatial jittering. However, classification improves even more when combining all three augmentation methods: we show the difference between each type of test-time augmentation and single-image classification accuracy in Fig.~\ref{fig:face_testaug}.

\myparagraph{Training with GAN-generated views.}
Due to the relative simplicity of the face domain and the GAN's ability to reasonably reconstruct aligned facial images, adding GAN-generated augmentations at test time is beneficial even when the classifier is not trained on the GAN domain. In the supplementary material, we investigate training the classifiers on GAN reconstructions (Eqn.~\ref{eqn:training_classifier}) and reconstructed latent code perturbations as additional augmentation.

\begin{figure}[t!]
  \centering
  \includegraphics[width=\linewidth]{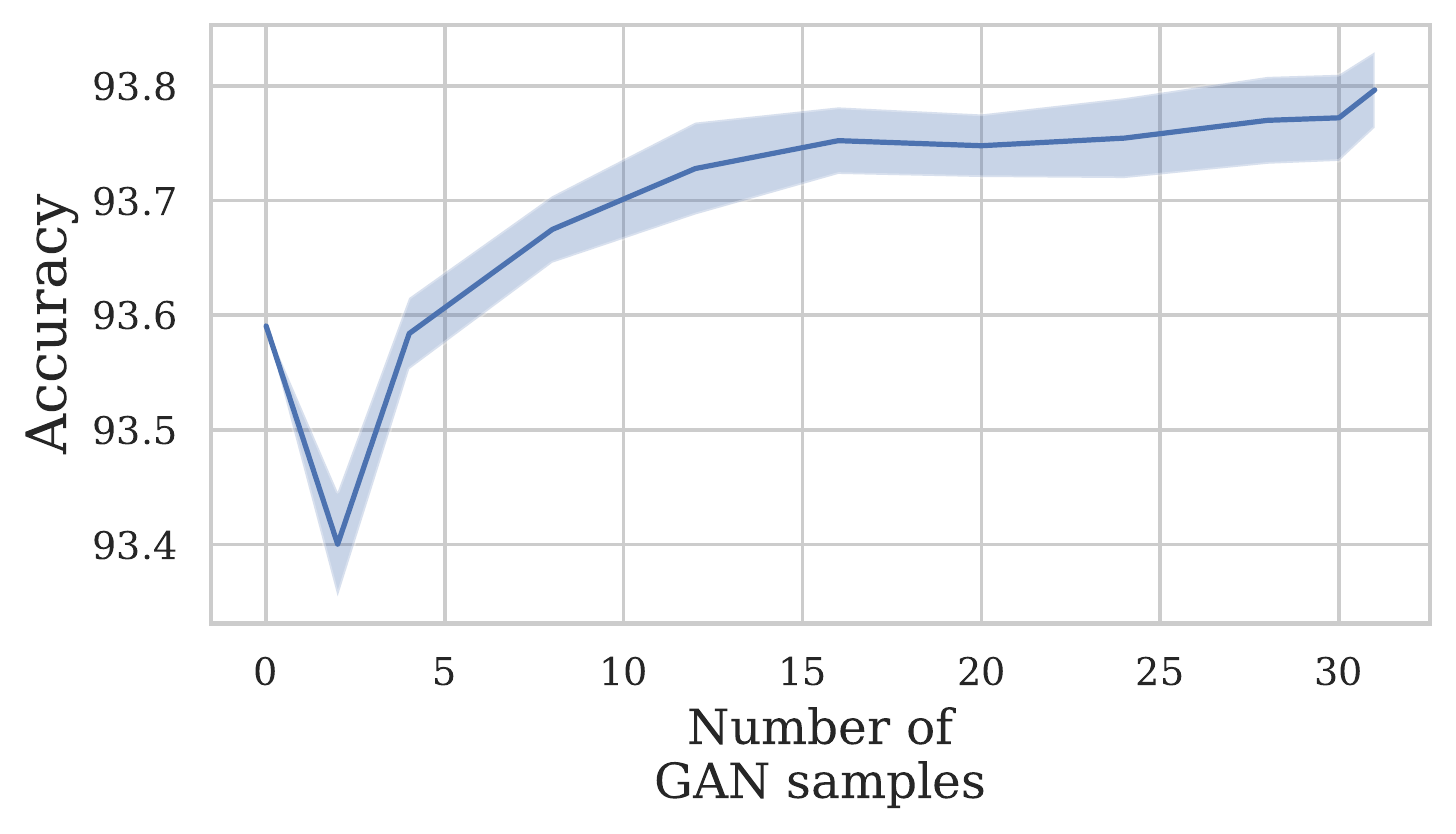}
  \vspace{-8mm}
  \caption{\small Classification accuracy as a function of the number of ensembled deep generative views on the Smiling attribute. Zero corresponds to using the original input image. Even adding a couple of views increases accuracy. Generally, adding more views further increases accuracy. We use ensemble size of 32 in our experiments, as performance saturates. We show additional attributes in the supplement. 
  \label{fig:ensemble_size}}
\end{figure}

\myparagraph{Investigating dependency on ensemble-size.}
While we use an ensemble size of 32 images (1 dataset image and 31 GAN views) for all prior experiments, here we investigate how the number of samples used in the ensemble impacts classification accuracy.
Using the Smiling classifier trained on dataset images (rather than GAN images), we plot accuracy as a function of the number of GAN samples in the ensemble (zero indicates only the real image is used for classification, corresponding to standard test accuracy) in Fig.~\ref{fig:ensemble_size}. We find that accuracy increases as the number of views used for classification increases but plateaus after a certain point, after which increasing the number of GAN samples has limited returns.%

\begin{figure}[t!]
  \centering
  \includegraphics[width=\linewidth]{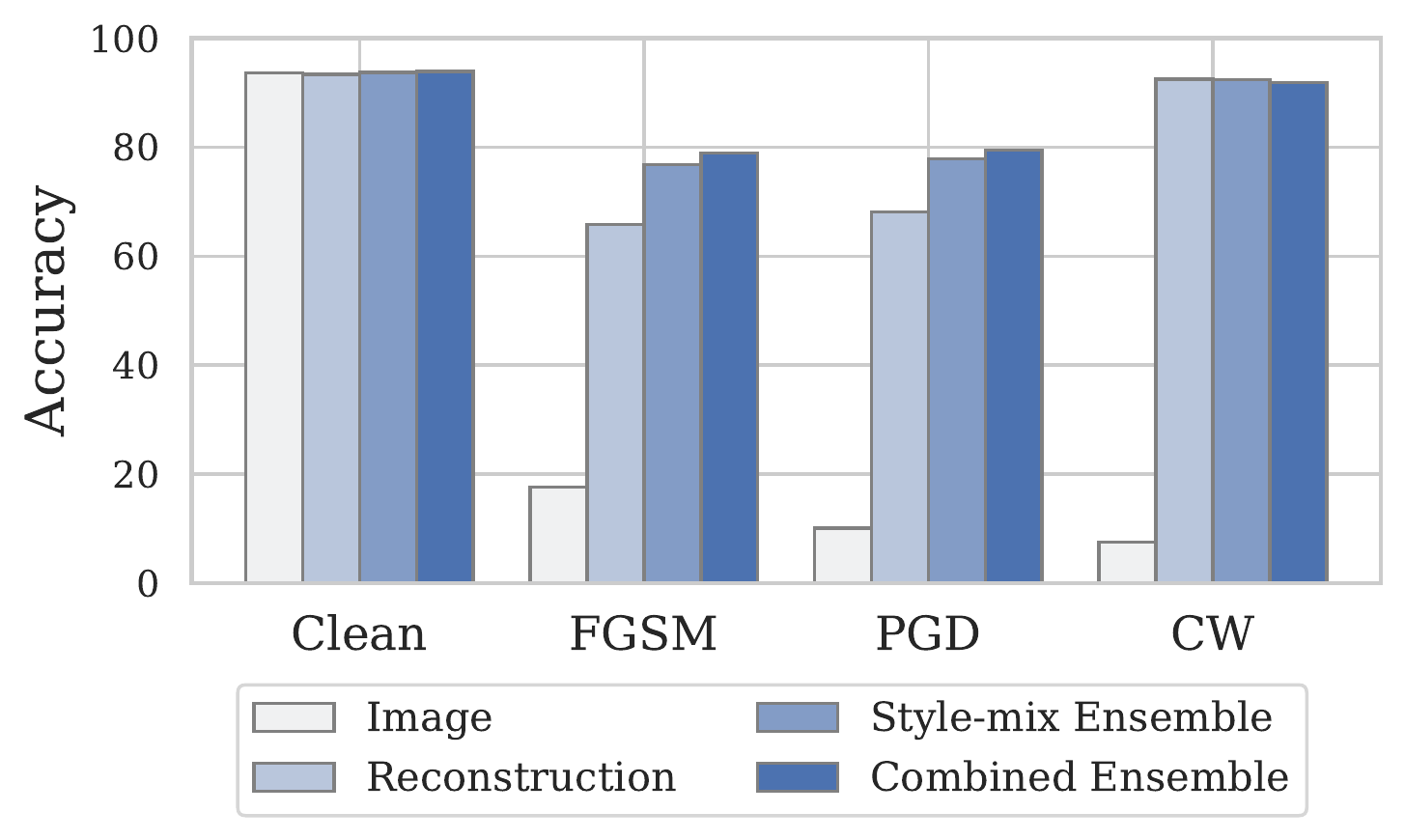}
  \vspace{-8mm}
  \caption{\small Robustness to adversarial attacks. We show accuracy on a corrupted image (Image), the GAN reconstruction (Reconstruction), ensembling with GAN style-mixing (Style-mix Ensemble), and ensembling over both traditional and GAN views (Combined Ensemble), on the Smiling attribute. Adversarial attacks (FGSM, PGD, CW) greatly reduce accuracy. On all cases, just GAN reconstruction recovers significant performance, and for FGSM and PGD attacks, ensembling with GAN views further improves accuracy. We show additional attributes and experiments on untargeted corruptions in the supplementary material.
  \label{fig:corruption}}
  \vspace{-6mm}
\end{figure}

\begin{figure*}[ht!]
  \centering
  \includegraphics[width=0.45\textwidth]{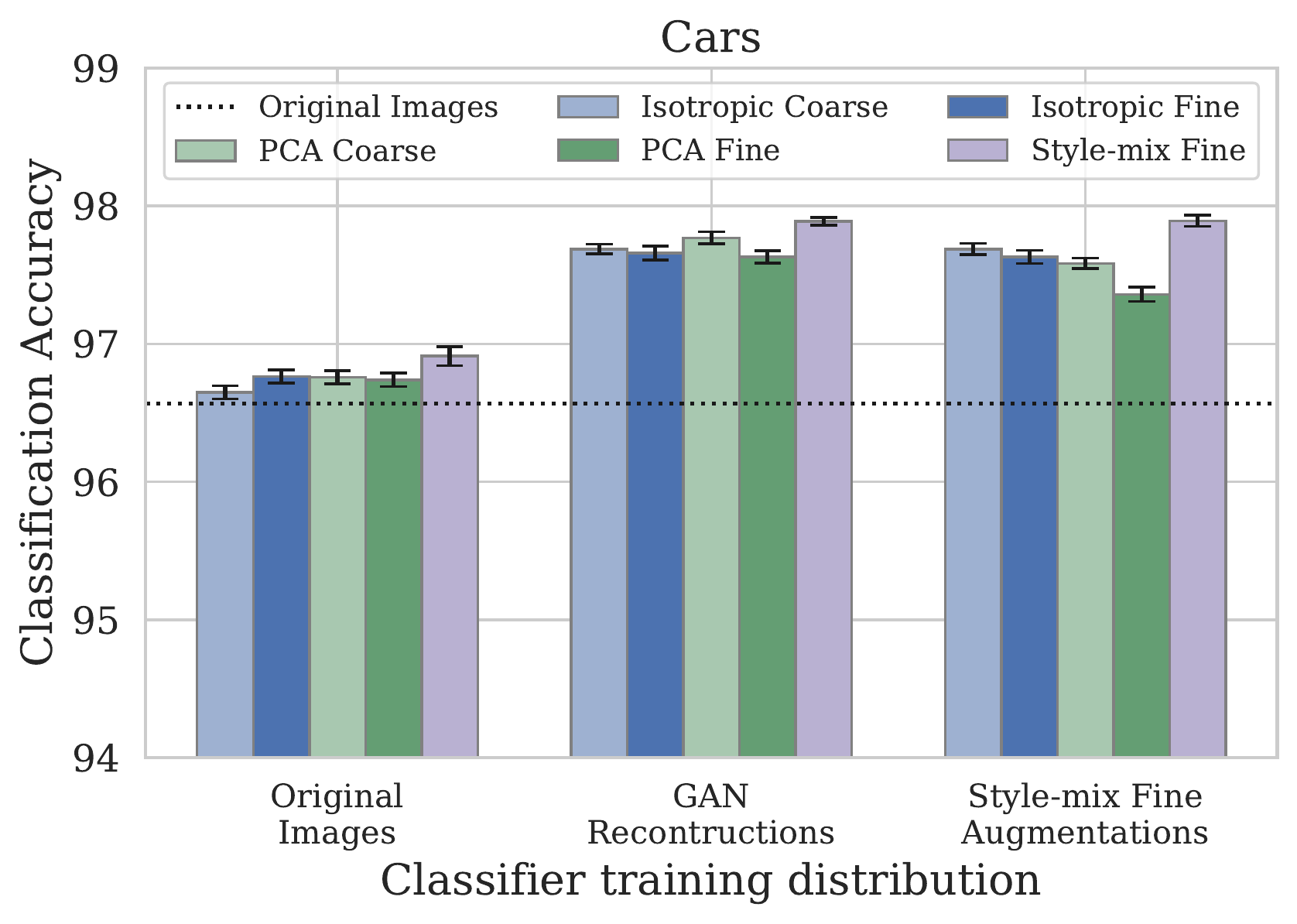}
  \hspace{0.2in}
  \includegraphics[width=0.45\textwidth]{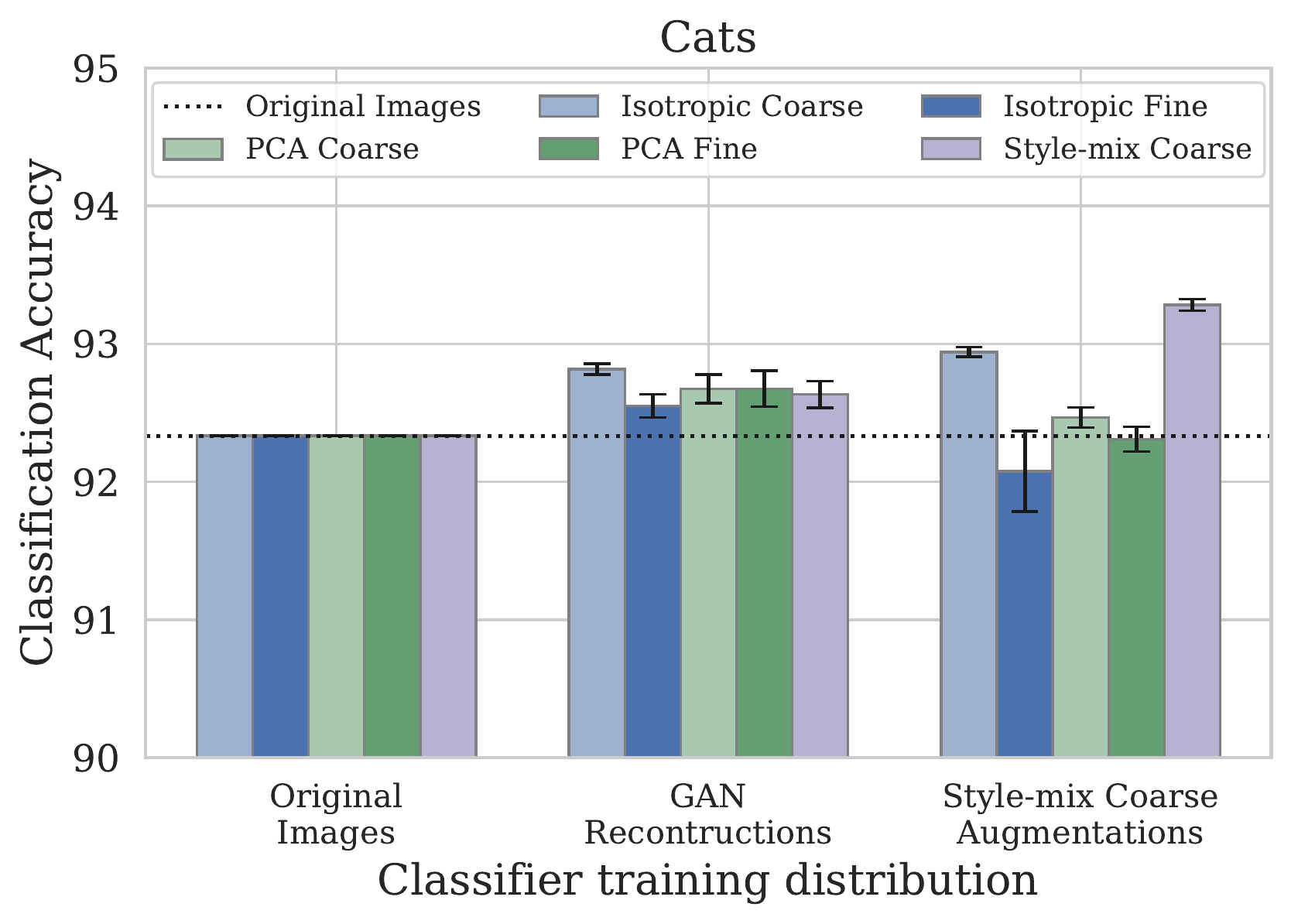}
  \vspace{-2mm}
  \caption{\small Effect of ensembling deep generative views. Classification accuracy for cars (left) and cats (right), as a function of training distribution (original images, GAN reconstructions, and GAN style-mixing, on x-axis) and test procedure (original images, and isotropic, PCA, and mixing-based GAN augmentations, as colored bars). 
  The dotted line is the baseline, trained and tested on real image crops. When the classifier is trained on the dataset images, GAN-generated views ensembled at test offers improvements only in the Car dataset.
  Fine-tuning the classifier on GAN-generated views improves the effects of test-time ensembling, and outperforms the baseline (error bars indicate standard error over bootstrapped samples from the ensemble). Fine layer style-mixing at test time is the best perturbation type on the cars domain, and coarse layer style-mixing is the best on the cats domain.
  The benefit of GAN augmentations on cat classification is weaker compared to cars; this is perhaps because cat classification is a harder problem (12-way vs. 3-way), resulting in a more constrained space of useful perturbations. We show results on additional classifier training distributions in the supplement.
  \label{fig:cat_car_graph}}
  \vspace{-2mm}
\end{figure*}

\myparagraph{Ensembling on corrupted images}
The GAN generator is trained only on clean images, and a projection step through the latent code can potentially map off-manifold images back to the image manifold, serving as a useful intermediary prior to a classification task.
Here, we investigate these scenarios using a set of targeted image corruptions, which adversarially change the output prediction of the classifier.
We test against FGSM~\cite{goodfellow2014explaining}, PGD~\cite{madry2017towards}, and CW attacks~\cite{carlini2017towards}. Our initial step, projecting the corrupted images through the latent code of the generator, corresponds to the DefenseGAN method~\cite{samangouei2018defense}. Classifying the corrupted images results in a large decrease in classification accuracy, which the projection step can partially recover. Next, we perturb the optimized latent code by applying style-mixing, and ensemble together these alternative views for classification. For the FGSM and PGD attacks, ensembled images from the GAN can boost classification beyond the single reconstructed images, and applying standard image augmentations in conjunction with the GAN-generated augmentations can offer an additional increase. In supplementary material we investigate targeted attacks on additional facial attributes and also untargeted corruptions, although we find that ensembling with GAN views offers stronger benefits in the targeted case compared to the untargeted scenario.

\subsection{Classifying Cars}

Due to its binary nature, the face classification problem is a relatively simple task, and given the success of recent GANs at imitating aligned faces, we find that adding views from the GAN at test time, without training on GAN-generated images, offers improvements. We next aim to investigate a slightly more difficult scenario of classifying types of cars, namely the `SUV', `Sedan', and `Cab' super-classes derived from \cite{krause2013object} (note that we use these super-classes, rather than the original 196 fine-grained classes, as the GAN reconstruction cannot recover fine details such as make and model of a car). Following best practices, we center the car using a bounding box prior to GAN projection, as GAN reconstruction tends to be better on centered objects~\cite{huh2020transforming}. After projecting the images into the GAN latent space, we try isotropic, PCA, or style-mixing pertubations  (Sec.~\ref{sec:method_perturbation}). Fig~\ref{fig:qualitative_perturbation} shows qualitative examples of fine and coarse style-mixing perturbations; the remaining perturbation methods are shown in supplementary material, along with additional dataset and classifier training details.

With a classifier trained on car images, ensembling the predicted classes over multiple random crops at test time boosts accuracy over a single crop (95.8\% vs. 96.6\% accuracy).
Adding additional GAN augmentations at test time offers a small benefit, with the fine style-mixing augmentation increasing accuracy to 96.9\%. 
However, due to imperfect reconstructions there can be a domain gap between the training domain of the classifier and the reconstructed images from the GAN. If we finetune this initial classifier using the GAN reconstructions or perturbed GAN reconstructions, the addition of GAN augmentations offers further benefits, increasing accuracy to 97.9\%. In Fig.~\ref{fig:cat_car_graph} we report classification accuracy on these training variations, 
and show results on additional training variations in supplementary material. For this domain, since using random image crops improves classification, we combine 16 random image crops and 16 random crops of the GAN-generated views at test time to form the ensemble; we average the classifier predictions on the random image crops with weight $1-\alpha$ and average the predictions on GAN outputs with weight $\alpha$ (Eqn.~\ref{eqn:ensemble_weight}). 

\subsection{Classifying Cats}

We next investigate a 12-way classification task on aligned cat faces from~\cite{parkhi2012cats}. %
Since the cat images are aligned, we find that ensembling with random image crops performs similarly to using a single center-cropped image, so we combine the real image with 31 random GAN-generated variants (32 images in total) following Eqn.~\ref{eqn:ensemble_weight}. 
As this task is more difficult than the previous two scenarios, we find that adding GAN augmentations does not improve accuracy when the classifier is only trained on real images. However, ensembling with GAN-generated views helps when the classifier is finetuned on the GAN domain. We note that training with the perturbed GAN reconstructions also benefits standard image classification as well, increasing from 92.3\% when trained only with images to 93.2\% when trained with GAN perturbations on the image dataset (shown in supplementary material), to which adding GAN augmentations at test time further increases to 93.3\%. We show additional training variations in the supplement. %

\section{Discussion}

While GANs show promise in synthesizing alternative views of a given image, several challenges remain in using them for downstream classification. Firstly, the GAN must be able to reconstruct the salient features used for classification. Here, we choose the StyleGAN2 generator~\cite{karras2019style}, which focus on modeling a single object category (such as faces, cars, and cats), rather than class-conditional models like BigGAN~\cite{brock2018large} on 1000 ImageNet categories~\cite{russakovsky2015imagenet}. Due to the higher variation in multi-class data, image projection (a critical part of generating deep augmentations) is still not sufficiently fast and reliable across the full dataset~\cite{huh2020transforming,pan2020exploiting,mao2020generative}. Even within a single-class StyleGAN2 generator, some aspects of the original image cannot be accurately recovered, such as fine textures, ornate backgrounds, or non-canonical poses (shown in supplementary material), impacting downstream classification.  

Related to GAN reconstruction \textit{quality} is the ability to \textit{efficiently} find a latent code that corresponds to a target image. Longer optimization can better reconstruct the image, but becomes intractable over a large dataset. In supplementary material, we investigate classification accuracy as a function of the number of optimization steps, and also perform experiments using an alternative inversion method~\cite{zhu2020domain} on a smaller face GAN, obtaining similar results. Moreover, recent alternative architectures trained specifically for efficiency~\cite{lin2021anycost} or invertibility~\cite{park2020swapping} may help further reduce the computational cost of image reconstruction.

On the classification side, we note that classifiers are sensitive to GAN reconstructions; accuracy on the reconstructions tends to be lower than that of the dataset images, requiring an ensemble weighting hyperparameter to merge the predictions of the image and the GAN outputs at test time. In most cases, we find that GAN transformations that modify style tend to be more beneficial than those that modify poses. This is in line with previous works that note the benefits of style-based \textit{training} augmentations for image classification~\cite{mao2020generative,cubuk2019autoaugment} and related positional sensitivities of classifiers~\cite{zhang2019making,azulay2018deep,engstrom2019exploring}. 
In the more difficult Imagenet classification problem, we found performance degrades substantially during image projection, and therefore GAN perturbations offer limited benefits. In supplementary material, we show results on using the class-conditional CIFAR10 StyleGAN2, where we also find the GAN reconstructions are more difficult to classify than the original images, such that adding GAN-reconstructed views does not benefit classification at both training and test time. When training classifiers, we use standard random flip, resize, and crop transformations on the image, but we further find that the alternative image augmentation strategies \textit{during training} can slightly outperform using the GAN-based augmentations \textit{at test time}, also shown in the supplement.

In our experiments, we find that ensembling images with GAN views helps in simple classification settings, but gains are small and are impacted by (1) the quality and efficiency of GAN reconstructions and (2) classifier sensitivities to the imperfect GAN outputs. Accordingly, as generative modeling technology advances in the future to create better reconstructions, similar GAN augmentation and ensembling strategies may yield even greater improvements. 

\section{Conclusion}

We investigate the capability of StyleGAN2 to generate alternative views of an image for use in downstream classification tasks. We first project the image into GAN latent space using a hybrid encoder and optimization setup to balance reconstruction speed and image similarity. Next, we investigate several types of perturbations in the StyleGAN2 latent space, such as isotropic Gaussian noise, Principal Component directions, or style-mixing the optimized latent code with a random latent code at certain generator layers.
We find that two adjustments to naive ensembling are beneficial: (1) appropriately weighting the real image and the GAN outputs in the ensembled predictions, and (2) finetuning the classifier on reprojected GAN samples to account for the domain gap between real images and GAN-generated variants. We conduct experiments on ensembling with GAN-generated perturbations on face attribute, cat, and car classification tasks, and investigate the impacts of ensemble size and corruptions in the input image. Due to current limitations in GAN reconstructions and classifier sensitivities, we are constrained to relatively simple tasks with small datasets, which may be mitigated with future improvements in generative modeling technology.

\myparagraph{Acknowledgments.} We would like to thank Jonas Wulff, David Bau, Minyoung Huh, Matt Fisher, Aaron Hertzmann, Connelly Barnes, and Evan Shelhamer for helpful discussions. LC is supported by the National Science Foundation Graduate Research Fellowship under Grant No. 1745302. This work was started while LC was an intern at Adobe Research.

{\small
\bibliographystyle{ieee_fullname}
\bibliography{egbib}
}

\clearpage
In supplementary material, we provide additional details on dataset preparation and classifier training methods for each classification task. We show additional qualitative examples of the GAN reconstructions and the perturbation methods investigated in the main text, at both fine and coarse layers of the latent code. Finally, we provide additional results investigating different experiment settings and classifier training distributions under each type of latent perturbation method.

\section{Supplementary Methods}

\subsection{Pretrained generators}

Unconditional GANs learn to mimic the image manifold by transforming low dimensional latent codes to image outputs. A number of interesting properties emerge in these generator networks, such as learning to model degrees of variation in real data. We use pretrained StyleGAN2 generators~\cite{karras2019analyzing} for our experiments. As class labels for images are not required during GAN training, the generators are trained on larger datasets than we would otherwise use for classification -- 5,520,756 images for LSUN Cars~\cite{yu15lsun}, 1,657,266 images for LSUN Cats~\cite{yu15lsun}, and 70,000 for FFHQ faces~\cite{karras2019style} for the 512$\times$384 resolution car, 256$\times$256 resolution cat, and 1024$\times$1024 resolution face generators respectively. 

\subsection{Datasets}

\paragraph{Face attribute classification}

We used the labeled CelebA-HQ~\cite{liu2018large,karras2017progressive} dataset containing 30000 faces with 40 labelled attributes. All face images are aligned by facial landmarks and cropped to square at 1024$\times$1024 resolution. We follow the training, validation, and test splits used in~\cite{liu2018large} on the 30,000 HQ images to obtain 24,183 images for training, 2993 for validation, and 2824 for testing. Due to the alignment and square cropping, we do not perform any additional resizing or shifting operations prior to projection into the GAN, i.e., the GAN reconstructs the full input image.

\paragraph{Car Shape Classification} We derive our cars dataset from ~\cite{krause2013object}, which in total contains 16,185 car images at the granularity of Make, Model, and Year for each image. As the GAN cannot recover fine-grained details for each image, instead we take a subset of the labelled images and group them into super-classes of ``SUV'', ``Sedan'', and ``Cab'' by parsing the provided class name. Using this subset of three super-classes, we divide the images into 2007 images for training, 1007 for validation, and 1049 for testing; by splitting each fine-level class according to a 50\%/25\%/25\% ratio. Prior to projection into the GAN's latent code, we first rescale the width of each image to 512px, shift the image to center the car using the provided bounding box, and then perform a center crop on the shifted image to fit the GAN's aspect ratio (512$\times$384 pixels). As the shifting operation may introduce unknown pixels around the edges of the image, the encoder and optimization step are both performed with a masking input to account for these missing pixels~\cite{chai2021latent}. Note that due to the GAN's aspect ratio and the aspect ratio of cars, there are parts of the image that may be cut out; therefore at test time, we find that adding multiple random image crops to ensemble improves classification.

\paragraph{Cat breeds classification}

The pets dataset from~\cite{parkhi2012cats} contains in total 37 breeds of cats and dogs, including 12 cat breeds with 200 images per class. We subdivide the 200 images of each class into 100 images for training, 50 for validation, and 50 for testing; this yields a total of 1200 images for training, 600 for validation and 600 for testing. To preprocess the dataset, we align each image using face attributes: we apply a face landmark detector on the images\footnote{\url{https://github.com/zylamarek/frederic}}, align the landmarks to a canonical pose, and crop to 256px; empirically we find that this improves both classification performance and GAN reconstruction. Note that in a few cases, the cat face is not correctly detected, resulting in a poorly aligned image; even though these instances will negatively impact classification, we do not remove them but rather retain the full dataset. Similar to before, for the GAN reconstruction process, the encoder and optimization steps both use a masking input to account for missing pixels that may occur during the alignment operation. 

\subsection{Classifier Training}

\paragraph{Binary facial attribute classification}

For classification of binary face attributes we follow the setup of Karras et al.~\cite{karras2019analyzing}, which uses the GAN discriminator architecture as the model for the attribute classification task. We also follow the corresponding downsampling step which performs a 4$\times$4 average pooling operation prior to classification. We train a classifier for each attribute from scratch (we also experimented with finetuning classifiers, and obtained similar results). Random horizontal flipping is applied during training. We use the Adam optimizer~\cite{kingma2014adam} with default parameters (learning rate $10^{-3}$, $\beta_1=0.9$), and train until validation accuracy does not increase for five epochs (most attributes will finish training by 20 epochs). We use the same setup when training from the image dataset as training from the GAN-generated reconstructions: depending on the setting, we simply replace the image $x$ with the reconstructed image $G(w^*)$, or the perturbed latent $G(\tilde{w})$ before sending to the classifier for training. We use the checkpoint with the highest validation accuracy for further experiments.

\paragraph{Multi-class classification} For the cat and car classification tasks, we use a ResNet-18~\cite{he2016deep} backbone with ImageNet~\cite{russakovsky2015imagenet} pretrained weights for the feature extractor. We modify the final linear layer to output the appropriate number of logits for each class -- three classes for car classification, and 12 classes for cat breeds. First, we finetune this model on the respective datasets. We use the Adam optimizer~\cite{kingma2014adam} with initial learning rate $10^{-4}$ for the feature backbone and $10^{-3}$ for the linear classification layer and $\beta_1=0.9$. We then decay learning rate by 10$\times$ if the validation accuracy does not increase for 10 epochs, up to a minimum learning rate of $10^{-6}$. We use a maximum of 500 epochs for training, and record the checkpoint with the highest validation accuracy for further experiments. Next, to finetune a model on GAN-generated samples, we start with this previous model finetuned on the appropriate dataset and train with a reduced learning rate $10^{-6}$; with 50\% probability for each batch, the model is finetuned on the GAN reconstructions or real image samples. During the classifier training procedure we apply a random resized crop with scale=[0.8, 1.0] and random horizontal flipping. All images are cropped to square prior to classification, at a resolution of 256$\times$256 for cars and 224$\times$224 for cats.

\subsection{Perturbations in GAN Latent Code}

For the isotropic and PCA direction perturbation methods, an additional hyperparameter is the extent of perturbation allowed. For the isotropic perturbation, we scale the variance $\sigma$ of the added noise to ensure that the modified latent does not deviate too far from the starting latent. For the PCA directions, we randomly sample a multiplier $\beta \sim \mathcal{U}[-\sigma, \sigma]$ that we use to scale the selected principle component direction. We try a few values for each hyperparameter, selected so that the GAN-modified outputs are similar to the input but with small distortions; we take the best setting from validation data to apply on the test partition. Values for these hyperparameters are listed in Table~\ref{tab:perturbations}.

\begin{table}[ht!]
  \caption{\small Hyperparameter values for isotropic and PCA latent perturbation methods. For the isotropic perturbation, we use a $\sigma$ hyperparameter to scale the variance of the random noise added to each optimized latent code. For the PCA perturbation method, we use $\sigma$ to denote a maximum magnitude for each principle component applied, and sample a magnitude $\beta$ randomly from $\beta \sim \mathcal{U}[-\sigma, \sigma]$. For each model, we select a few discrete values for each hyperparameter to create small variations on the input image without deviating too far from the input. We select the best hyperparameter setting on validation data and use the same value at test time. \label{tab:perturbations}}
  \centering
  \resizebox{1.0\linewidth}{!}{
  \begin{tabular}{ccccc}
    \toprule
    \multirow{2}{*}{\bf Model} &
    \multirow{2}{*}{\shortstack[c]{\bf Isotropic\\\textbf{Coarse}}} &
    \multirow{2}{*}{\shortstack[c]{\bf Isotropic\\\textbf{Fine}}} &
    \multirow{2}{*}{\shortstack[c]{\bf PCA\\\textbf{Coarse}}} &
    \multirow{2}{*}{\shortstack[c]{\bf PCA\\\textbf{Fine}}} \\
    & & & & \\ %
    \midrule
    Car & \{1.0, 1.5, 2.0\} & \{0.3, 0.5, 0.7\} & \{1.0, 2.0, 3.0\} & \{ 1.0, 2.0, 3.0\} \\
    Cat &\{0.5, 0.7, 1.0\} & \{0.1, 0.2, 0.3\} & \{0.5, 0.7, 1.0\} & \{0.5, 0.7, 1.0\} \\
    Face & \{0.1, 0.2, 0.3\} & \{0.1, 0.2, 0.3\} & \{1.0, 2.0, 3.0\} & \{1.0, 2.0, 3.0\} \\
    \bottomrule
  \end{tabular}
  }
\end{table}

\section{Supplementary Results}

\subsection{Additional Qualitative Examples}

\begin{figure*}[ht!]
  \centering
  \begin{subfigure}[t]{1.0\textwidth}
    \centering
    \includegraphics[width=0.95\textwidth]{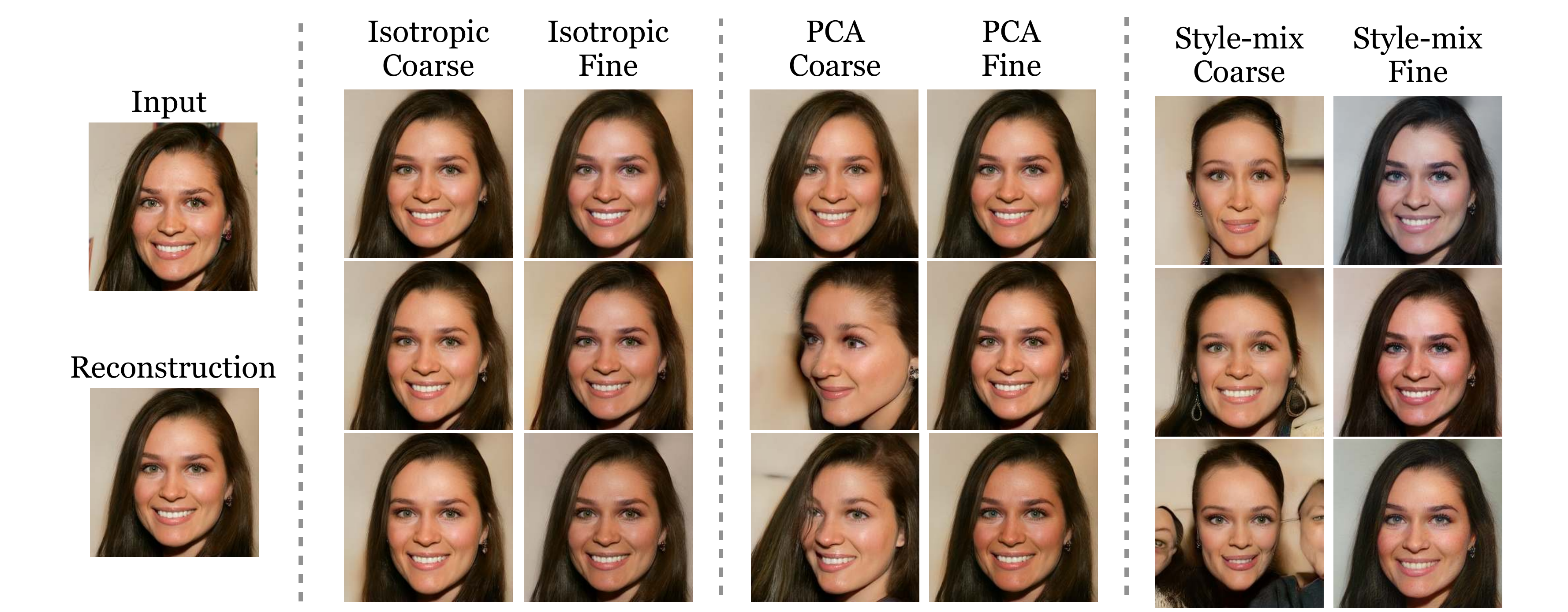}
    \caption{\small Face domain}
  \end{subfigure}
  \begin{subfigure}[t]{1.0\textwidth}
    \centering
    \includegraphics[width=0.95\textwidth]{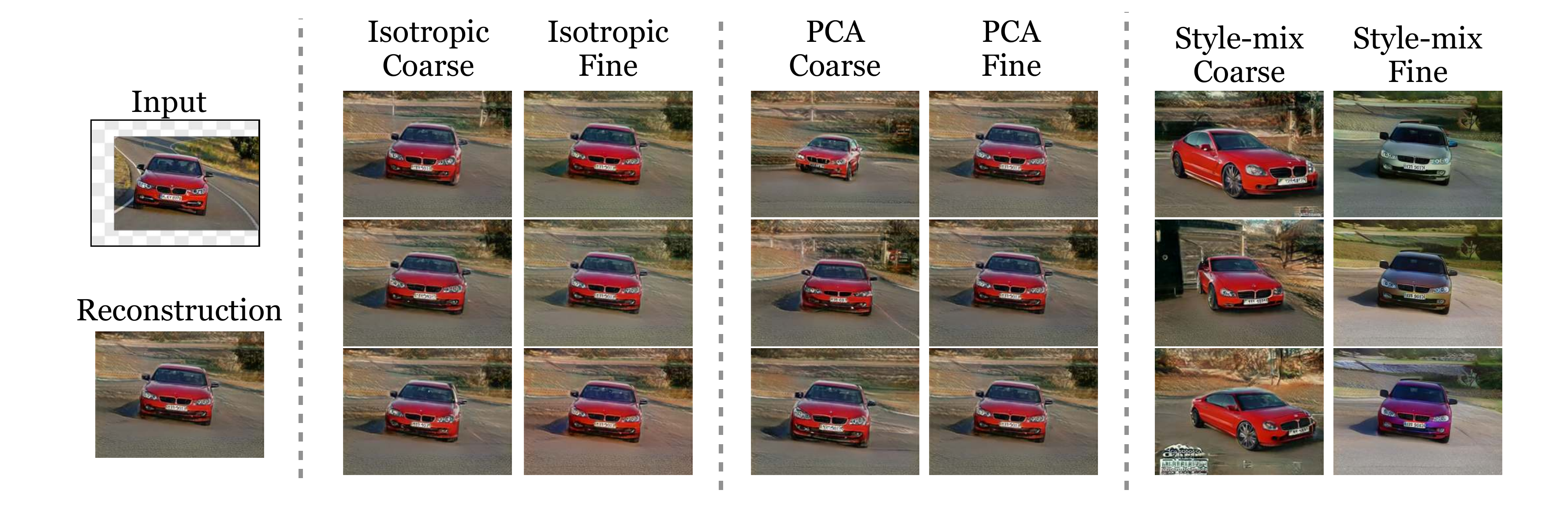}
    \caption{\small Car domain}
  \end{subfigure}
  \begin{subfigure}[t]{1.0\textwidth}
    \centering
    \includegraphics[width=0.95\textwidth]{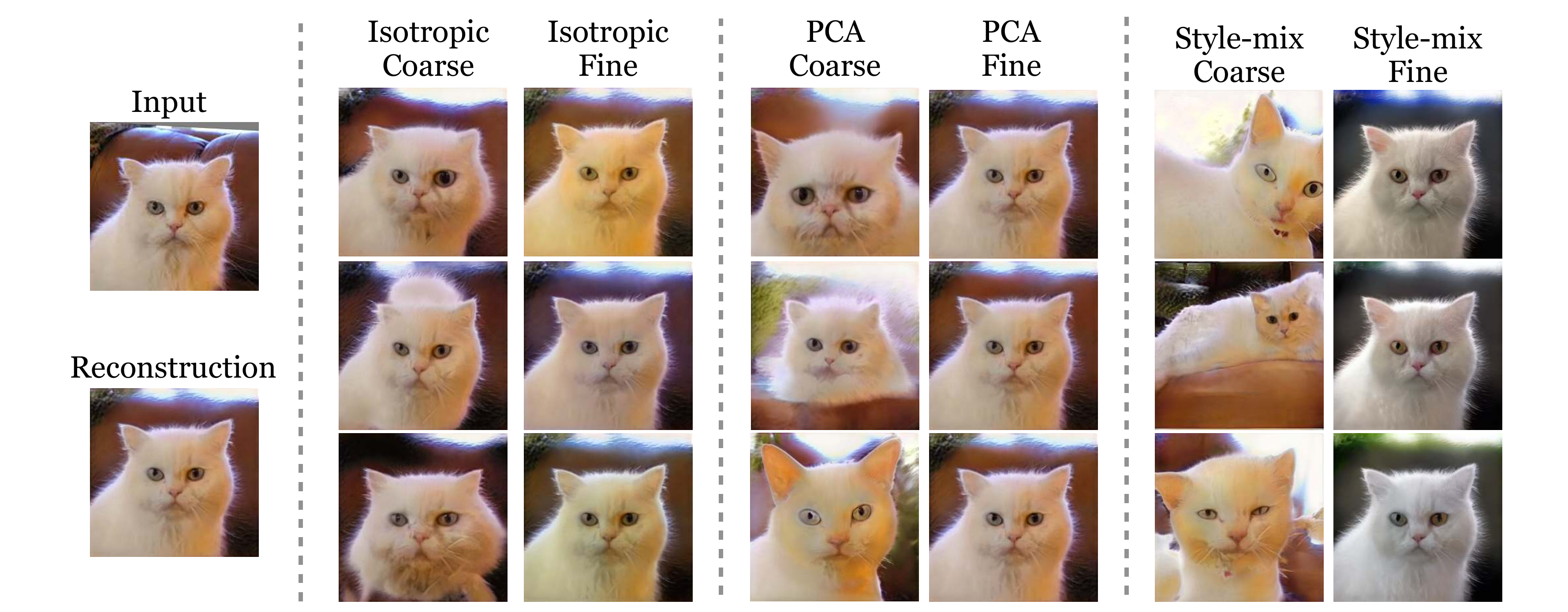}
    \caption{\small Cat domain}
  \end{subfigure}
  \caption{\small Visualizing GAN perturbations. For the (a) Face, (b) Car, and (c) Cat domains, we show qualitative samples of an input image (Input), which is centered, if necessary, prior to reconstruction by the GAN (Reconstruction). Once the latent code is optimized to obtained the best reconstruction of the input, we perform three types of latent code modifications: isotropic, PCA, and style-mixing perturbations at both coarse and fine layers of the latent code, where coarse-level manipulations alter pose and size, while fine-level manipulations alter color. 
  \label{fig:sm_qualitative}}
  \vspace{-0.2in}
\end{figure*}

In the main text, we define three methods for perturbations in the latent code of a GAN: 1) adding isotropic Gaussian noise, 2) moving along principle component axes~\cite{harkonen2020ganspace}, and 3) style-mixing the optimized latent code with a random latent code. We apply each type of perturbation respectively to the coarse layers (first four style layers) or fine layers (tenth and higher style layers) of the optimized latent code $w$. In Fig.~\ref{fig:sm_qualitative}, we show qualitative examples of each type of perturbation applied to the same base image in each domain. Note that the coarse layers correspond to positional or shape changes, while the fine layers correspond to coloring changes. Furthermore, the style-mixing operation, which swaps in an entirely \textit{random} latent code rather than adding some offset to the optimized latent, achieves qualitatively larger changes than the isotropic or PCA methods. 

\begin{figure*}[ht!]
  \centering
  \includegraphics[width=\textwidth]{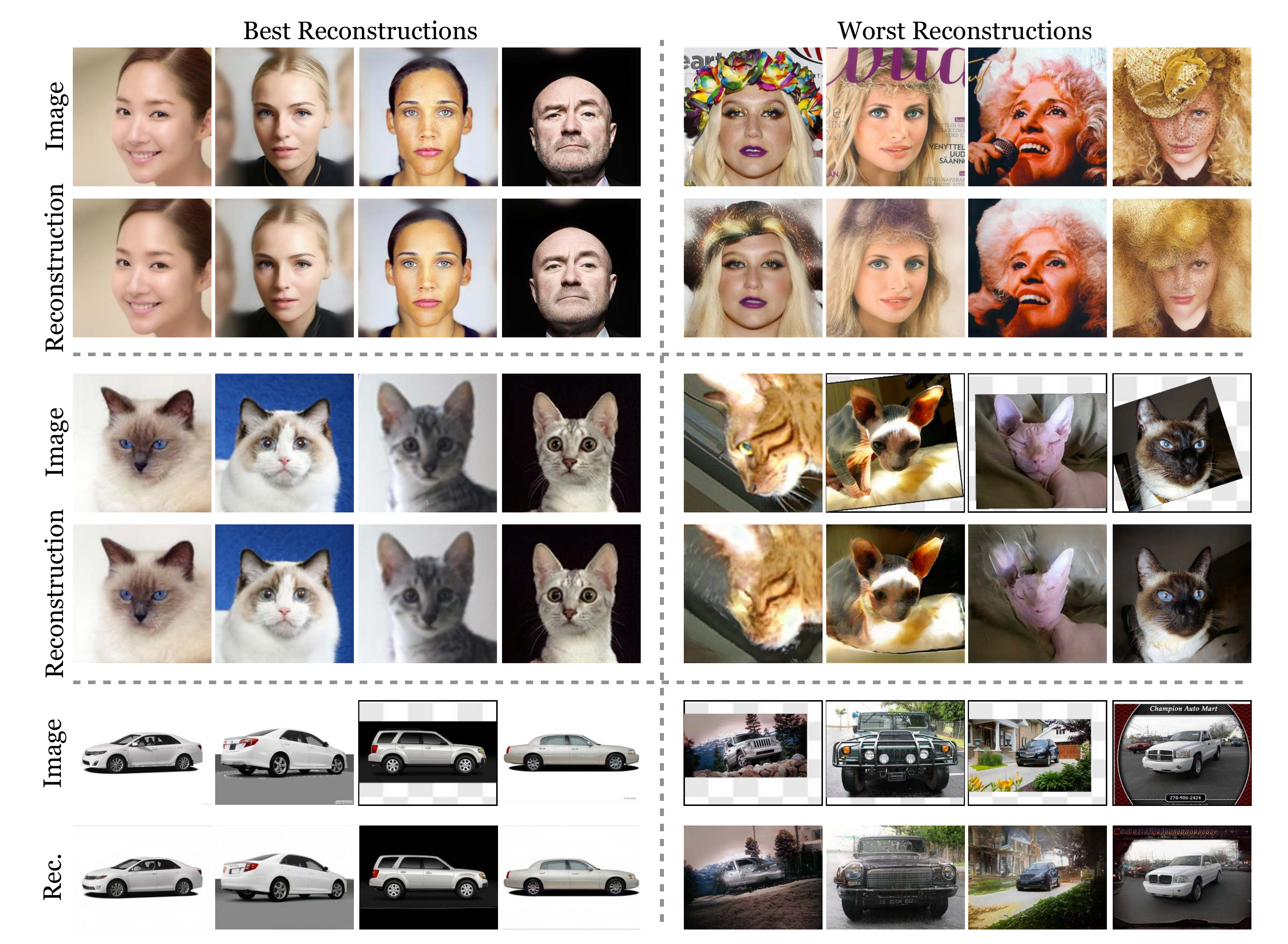}
  \caption{\small Visualizing the four best and worst reconstructions in the test split measured using LPIPS perceptual distance ~\cite{zhang2018unreasonable}. On each domain, The best reconstructions tend to be in canonical poses with simple backgrounds, and the worst reconstructions have complex backgrounds or textural details that the GAN cannot accurately recreate. 
  \label{fig:sm_reconstructions}} 
\end{figure*}

\begin{figure*}[ht!]
  \centering
\includegraphics[width=\textwidth]{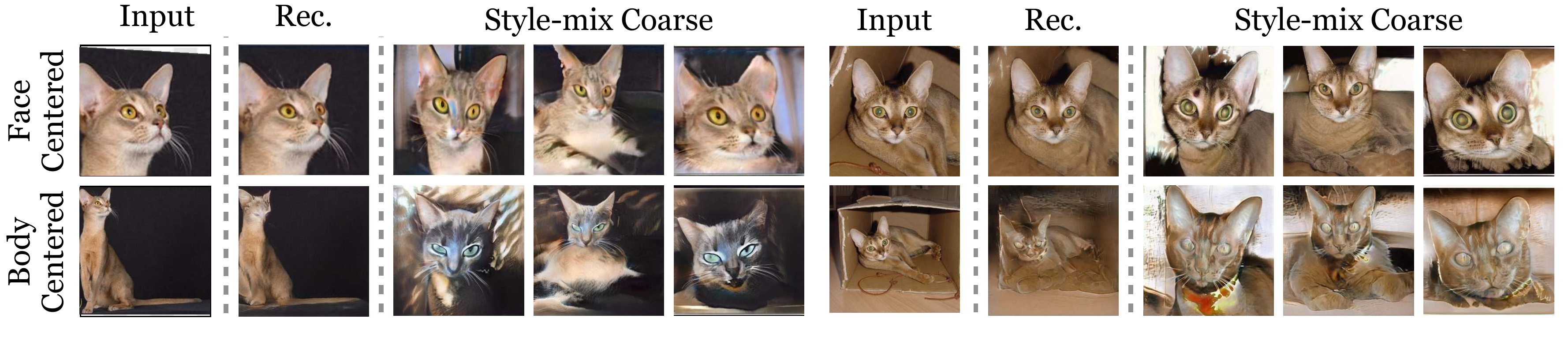}
  \caption{\small The StyleGAN2 generator exhibits a bias towards cat faces, thus we find that projecting face-centered images using the GAN yields better reconstructions than body-centered images. Furthermore, when the style-mixing operation is performed, the face-centered images better preserve the identity of the cat on the modified images.  
  \label{fig:sm_cat_face}}
\end{figure*}

Reconstruction via GAN inversion is easier for images in canonical poses and plain backgrounds that do not contain uncommonly seen details. Fig.~\ref{fig:sm_reconstructions} visualizes the four best reconstructed  and the four worst reconstruction images in the test split of each dataset, measured using the LPIPS perceptual distance metric~\cite{zhang2018unreasonable}. The hardest images to reconstruct contain text, large accessories on the head, or non-facial objects for the face domain. On the cat and car domains, the difficult cases are detailed textures, complex backgrounds, or unusual poses.

\begin{figure*}[ht!]
 \centering
  \includegraphics[width=\textwidth]{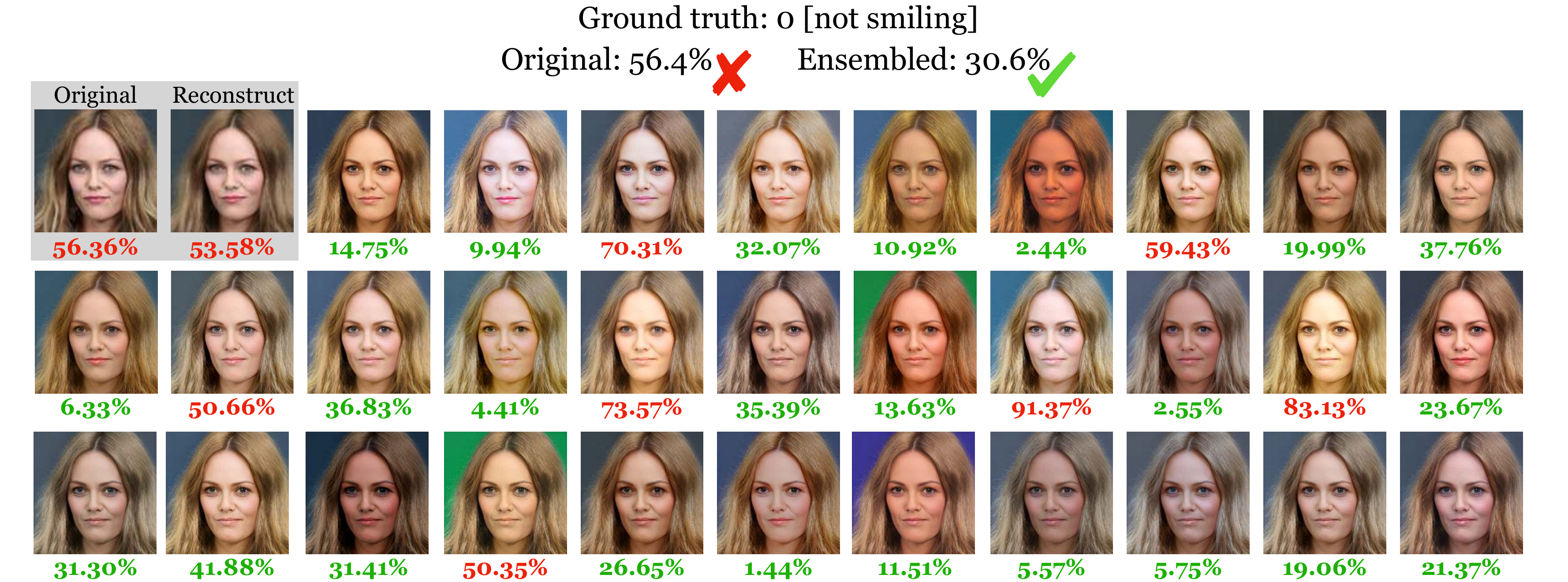}
  \caption{\small A selected example in which the original image is predicted incorrectly ($P(\mathrm{Smiling}) > 0.5$), but ensembling the classifier predictions with style-mixing on fine layers recovers the ground-truth label ($P(\mathrm{Smiling}) < 0.5$).
  \label{fig:sm_face_ensemble}}
\end{figure*}

On the cat dataset, we preprocess all images by aligning and cropping the face, as we find that the GAN has a facial bias in reconstruction. We show examples in Fig.~\ref{fig:sm_cat_face}. Centering the same image on the face, rather than the body (we use a MaskRCNN object detector~\cite{he2017mask} to obtain a bounding box for the cat), improves the GAN's reconstruction. Furthermore, we find that style-mixing in coarse layers better preserves the identity of the cat when it is face-centered, as opposed to body-centered.

In Fig.~\ref{fig:sm_face_ensemble} we show an example of the fine layer style-mixing augmentation in latent space where the original image is misclassified but the ensemble recovers the correct prediction. The classifier is sensitive to the variations introduced by the style-mixing operation, which changes the classifier's incorrect prediction of the original image to the correct one when averaging the predictions on the GAN-generated views.

\subsection{Additional Experiments: CelebA-HQ}

\begin{figure}[t!]
  \centering
  \includegraphics[width=\linewidth]{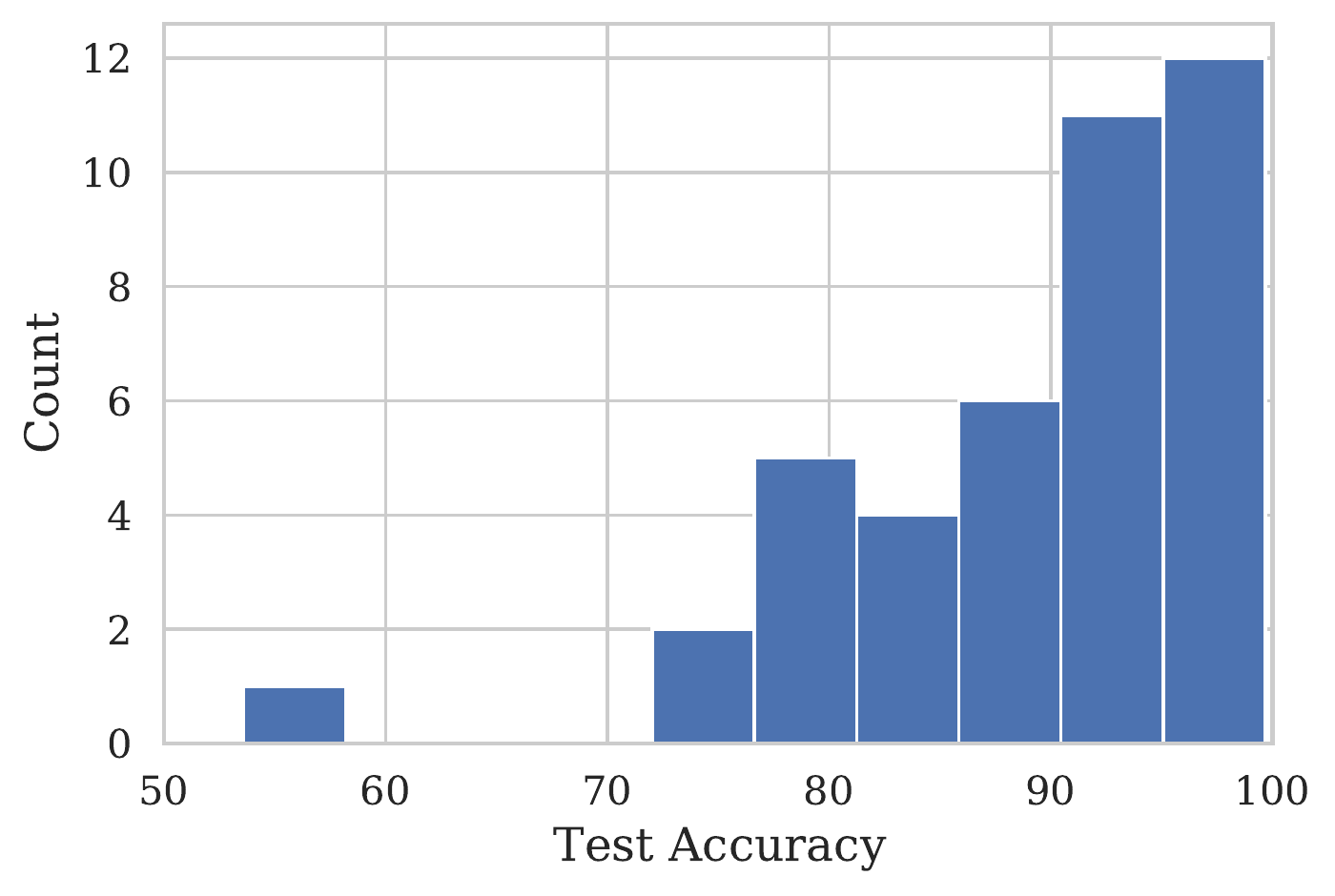}
  \caption{\small Distribution of test accuracies for the 40 attribute classification tasks in the CelebA-HQ dataset. Most attributes (23 out of 40) attain over 90\% accuracy on the test partition, while the Big Lips attributes has the test lowest accuracy at  53.58\%.
  \label{fig:sm_faces_distribution}}
\end{figure}

\paragraph{Distribution of classification accuracies} The CelebA-HQ dataset~\cite{liu2018large} contains attribute labels for 40 binary classification tasks spanning a wide range of difficulty. The hardest attribute to classify is ``Big Lips'' with a test of 53.58\%, while 23 out of 40 attributes can attain a test accuracy of over 90\%. We show a distribution of the test accuracy over all 40 attributes in Fig.~\ref{fig:sm_faces_distribution}. 

\begin{figure}[t!]
  \centering
  \begin{subfigure}[t]{1.0\linewidth}
    \centering
    \includegraphics[width=0.95\linewidth]{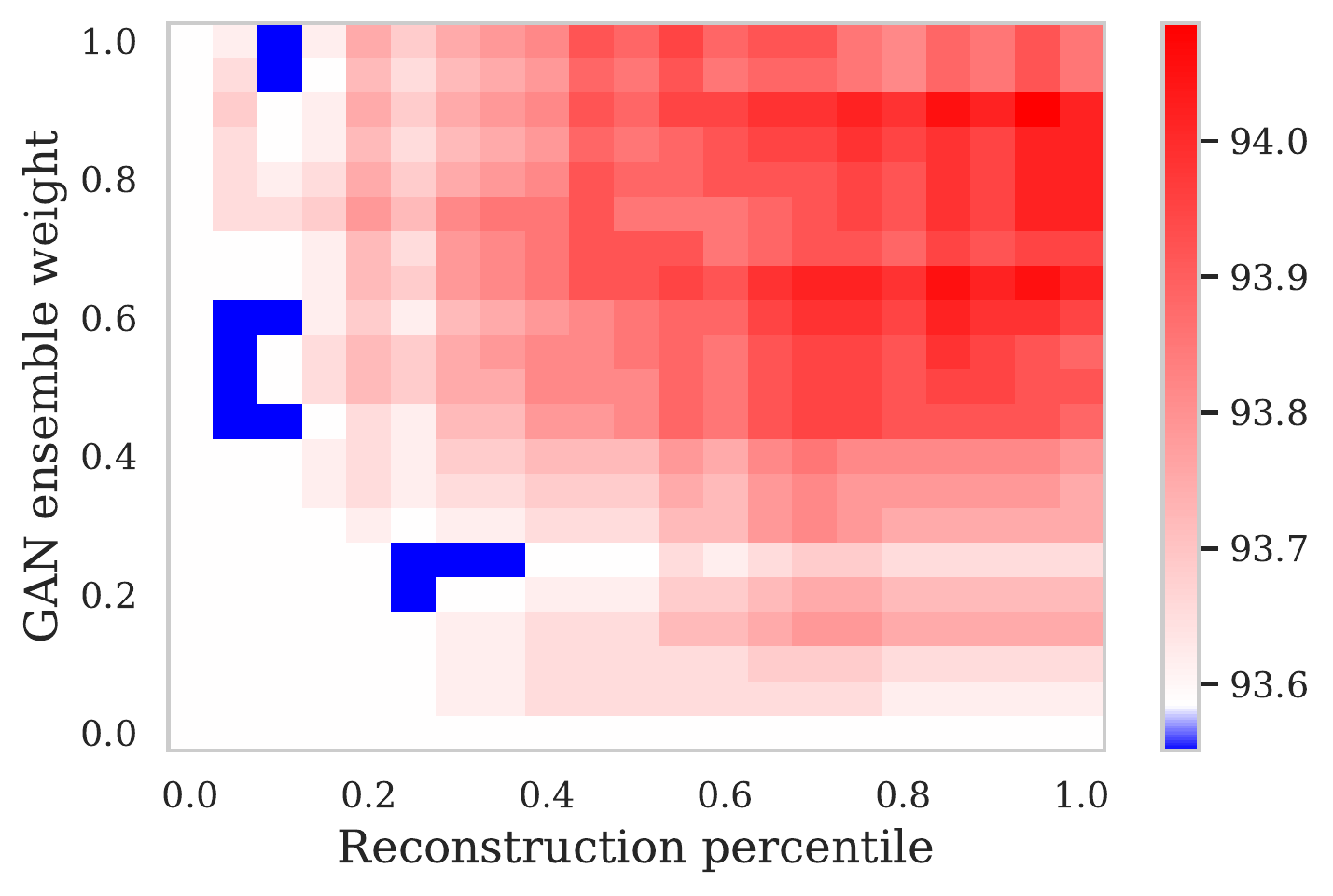}
    \vspace{-3mm}
    \caption{\small Smiling}
  \end{subfigure}
  \begin{subfigure}[t]{1.0\linewidth}
    \centering
    \includegraphics[width=0.95\linewidth]{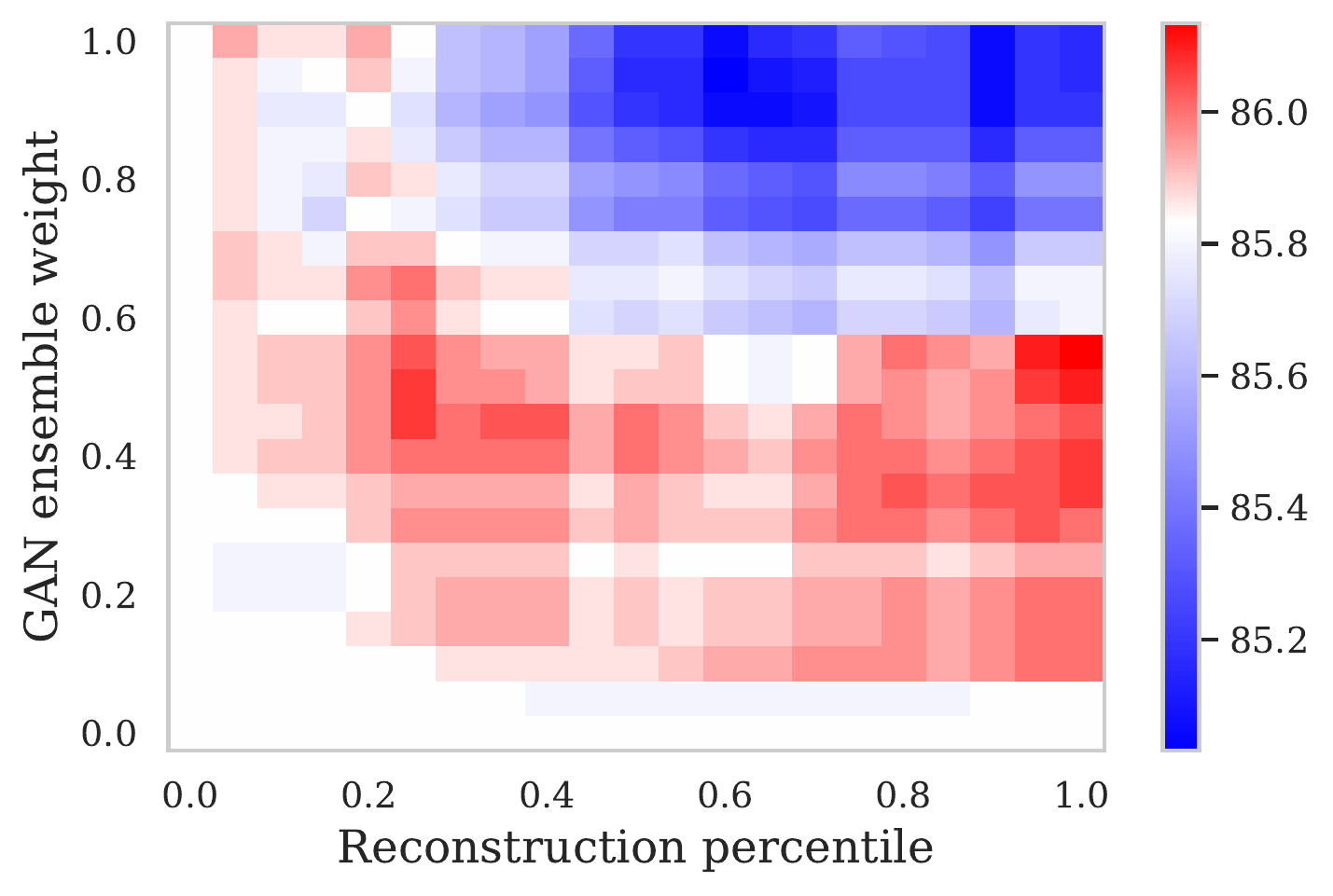}
    \vspace{-3mm}
    \caption{\small Young}
  \end{subfigure}
  \caption{\small Visualizing classification accuracy as a function of ensemble weight and reconstruction similarity. In the main text, we cross validate for an ensembling weighting parameter $\alpha$ between the original test image, and it's GAN-reconstructed variants. Here, we also explore reconstruction quality as an additional axis: we discard the GAN-reconstructed ensemble if the reconstruction similarity of the image is below a certain cutoff (0 corresponds to using no reconstructed images, and 1 indicates using the GAN ensemble for all images; the experiments in the main text correspond to the right-most column of each grid.) White corresponds to standard classification accuracy, shades of red indicate increases in accuracy, and shades of blue indicate decreases in accuracy. We find that the classification is more sensitive to the ensemble weighting $\alpha$ rather than reconstruction similarity, and using the 2D grid search performs similarly to the simpler 1D search over ensemble weight $\alpha$.
  \label{fig:sm_2d_grid}}
  \vspace{-0.2in}
\end{figure}

\paragraph{Variations on ensemble weighting} In the main text, we introduce a weighting hyper-parameter $\alpha$ which balances between using the original image for classification and the GAN-generated variants. However, not all dataset images can be reconstructed with the same fidelity. We investigate an alternative ensembling approach, in which we also discard the GAN-generated variants whose reconstruction error is greater than a certain percentile cutoff; this 2D space is visualized in Fig. ~\ref{fig:sm_2d_grid}. In the main paper, we retain the GAN-generated variants on all images and only use the ensemble weighting $\alpha$, which corresponds to a search over the right-most column in the 2D plot. However, we find that using the 2D search over ensemble weighting and reconstruction on the validation split performs similarly to the simpler 1D hyper-parameter search.

\paragraph{Optimization time vs. accuracy} For the classifier to behave similarly on the GAN reconstruction and a given real image, we desire the GAN's version to be as similar to the original image as possible. However, obtaining a close reconstruction via optimization is slow; we must balance between reconstruction quality and a computationally feasible optimization budget over the dataset. As such we use an encoder model to initialize the starting latent code, and then optimize for 500 steps to improve the reconstruction. In Tab.~\ref{tab:optimization}, we compare the L1 and LPIPS reconstruction distance and classification accuracy of the reconstructed images as a function of the number of optimization steps. While the reconstruction improves as more optimization is performed, the accuracy plateaus after 250 optimization steps, suggesting that a reduction in optimization time can possible while obtaining similar classification results.

\begin{table}[h!]
  \caption{\small Reconstruction similarity (L1, LPIPS) and  accuracy for the `Smiling' attribute, vs. optimization steps. Classification accuracy on images is 93.6\%.\label{tab:optimization}}
  \vspace{-1em}
  \centering
  \resizebox{1.0\linewidth}{!}{
    \begin{tabular}{llllllll}
     Opt. Steps& 0 & 50 & 100 & 150 & 200 & 250 & 500 \\
     \hline
     L1  & 0.172 & 0.104 & 0.092 & 0.086 & 0.083 & 0.080 & \bf 0.073 \\
     LPIPS & 0.443 & 0.252 & 0.219  & 0.201 & 0.188 & 0.179 &  \bf 0.152 \\
     Acc & 92.2 & 93.2 & 93.2 & 93.2 &  93.5 & \bf 93.6 & 93.4 \\
    \end{tabular}
}
\end{table}

\paragraph{Ensembled classification accuracy: 40 attributes}

When training the classifier, the standard approach is to train on the image dataset -- in this case, if GAN-generated images are added as part of the ensemble at test time, there is potentially a domain gap as the classifier has never seen GAN images during training. However, the face domain is fairly simple for the generator to reconstruct; adding fine layer style-mixing of images at test time, even without a classifier trained on GAN images, outperforms the baseline of testing on a single image when averaged over 40 attributes. Adding additional color and spatial jitter to the image at test time offers a further boost. Using this setting, we sort the attributes based on how much ensembling at test time helps in Table~\ref{tab:ensemble_diff}; the highest difference between ensembled test accuracy and standard test accuracy (classifying a single image) indicates the attribute where ensembling increases accuracy the most, while the lowest difference indicates where ensembling helps the least and can harm classification. There are a few attributes that notably do not benefit from ensembling: some accessories (Wearing Hat, Wearing Necklace) which can be difficult for the GAN to reconstruct, while several color-based attributes (Black Hair, Brown Hair, Gray Hair) also do not benefit from the fine layer style-mixing operation, which can cause color changes.

\begin{table*}[ht!]
  \caption{\small Comparison of standard test accuracy and ensembled test accuracy on classifiers trained on images, using the combined GAN augmentation at test time on 40 facial classification attributes. This augmentation consists of style-mixing at fine layers, and small color and spatial jittering. Attributes are sorted in order from highest difference (ensembling helps the most), to lowest difference (ensembling harms classification). 
  \label{tab:ensemble_diff}}
  \centering
  \resizebox{1.0\linewidth}{!}{
  \begin{tabular}{lccc | lccc}
    \toprule
    \multirow{2}{*}{\bf Attribute} &
    \multirow{2}{*}{\shortstack[c]{\bf Standard Test\\\textbf{Accuracy}}} &
    \multirow{2}{*}{\shortstack[c]{\bf Ensembled\\\textbf{Test Accuracy}}} &
    \multirow{2}{*}{\shortstack[c]{\bf Difference}} & 
    \multirow{2}{*}{\bf Attribute} &
    \multirow{2}{*}{\shortstack[c]{\bf Standard Test\\\textbf{Accuracy}}} &
    \multirow{2}{*}{\shortstack[c]{\bf Ensembled\\\textbf{Test Accuracy}}} &
    \multirow{2}{*}{\shortstack[c]{\bf Difference}} \\ 
    & & & & & & & \\ %
    \midrule
    1: Wearing Lipstick     & 93.17 & 94.47 & 1.30
     & 21: Wearing Earrings     & 85.20 & 85.21 & 0.02 \\
    2: Wavy Hair            & 74.29 & 75.26 & 0.96
     & 22: Bald                 & 98.26 & 98.28 & 0.01 \\
    3: High Cheekbones      & 85.48 & 86.40 & 0.92
     & 23: Mustache             & 95.89 & 95.90 & 0.00 \\
    4: No Beard             & 94.83 & 95.47 & 0.64
     & 24: Blurry               & 99.68 & 99.68 & 0.00 \\
    5: Goatee               & 95.82 & 96.23 & 0.41
     & 25: Bushy Eyebrows       & 91.50 & 91.50 & -0.00 \\
    6: Arched Eyebrows      & 81.59 & 81.95 & 0.36
     & 26: Double Chin          & 94.48 & 94.47 & -0.00 \\
    7: Male                 & 97.31 & 97.65 & 0.34
     & 27: Attractive           & 78.82 & 78.82 & -0.01 \\
    8: Mouth Slightly Open  & 93.52 & 93.83 & 0.31
     & 28: Chubby               & 94.37 & 94.36 & -0.01 \\
    9: Smiling              & 93.59 & 93.89 & 0.30
     & 29: Pale Skin            & 97.17 & 97.15 & -0.01 \\
    10: Young                & 87.50 & 87.74 & 0.24
     & 30: Rosy Cheeks          & 91.64 & 91.61 & -0.03 \\
    11: Eyeglasses           & 99.29 & 99.50 & 0.21
     & 31: Bangs                & 95.15 & 95.11 & -0.04 \\
    12: Bags Under Eyes      & 81.98 & 82.15 & 0.17
     & 32: Pointy Nose          & 72.24 & 72.18 & -0.06 \\
    13: Sideburns            & 96.67 & 96.83 & 0.16
     & 33: Gray Hair            & 98.37 & 98.28 & -0.09 \\
    14: 5 o Clock Shadow     & 92.71 & 92.84 & 0.13
     & 34: Receding Hairline    & 92.63 & 92.55 & -0.09 \\
    15: Big Nose             & 76.95 & 77.08 & 0.13
     & 35: Wearing Necklace     & 81.02 & 80.92 & -0.10 \\
    16: Heavy Makeup         & 89.09 & 89.22 & 0.13
     & 36: Brown Hair           & 86.61 & 86.49 & -0.13 \\
    17: Blond Hair           & 94.30 & 94.37 & 0.07
     & 37: Narrow Eyes          & 86.15 & 86.02 & -0.13 \\
    18: Oval Face            & 79.32 & 79.36 & 0.04
     & 38: Black Hair           & 89.41 & 89.14 & -0.27 \\
    19: Wearing Necktie      & 95.47 & 95.50 & 0.04
     & 39: Wearing Hat          & 98.58 & 98.29 & -0.29 \\
    20: Straight Hair        & 80.38 & 80.42 & 0.04
     & 40: Big Lips             & 53.58 & 53.05 & -0.52 \\
    \bottomrule
  \end{tabular}
  }
\end{table*}

\paragraph{Alternative projection algorithms}

\begin{figure}[t!]
  \centering
  \includegraphics[width=0.45\textwidth]{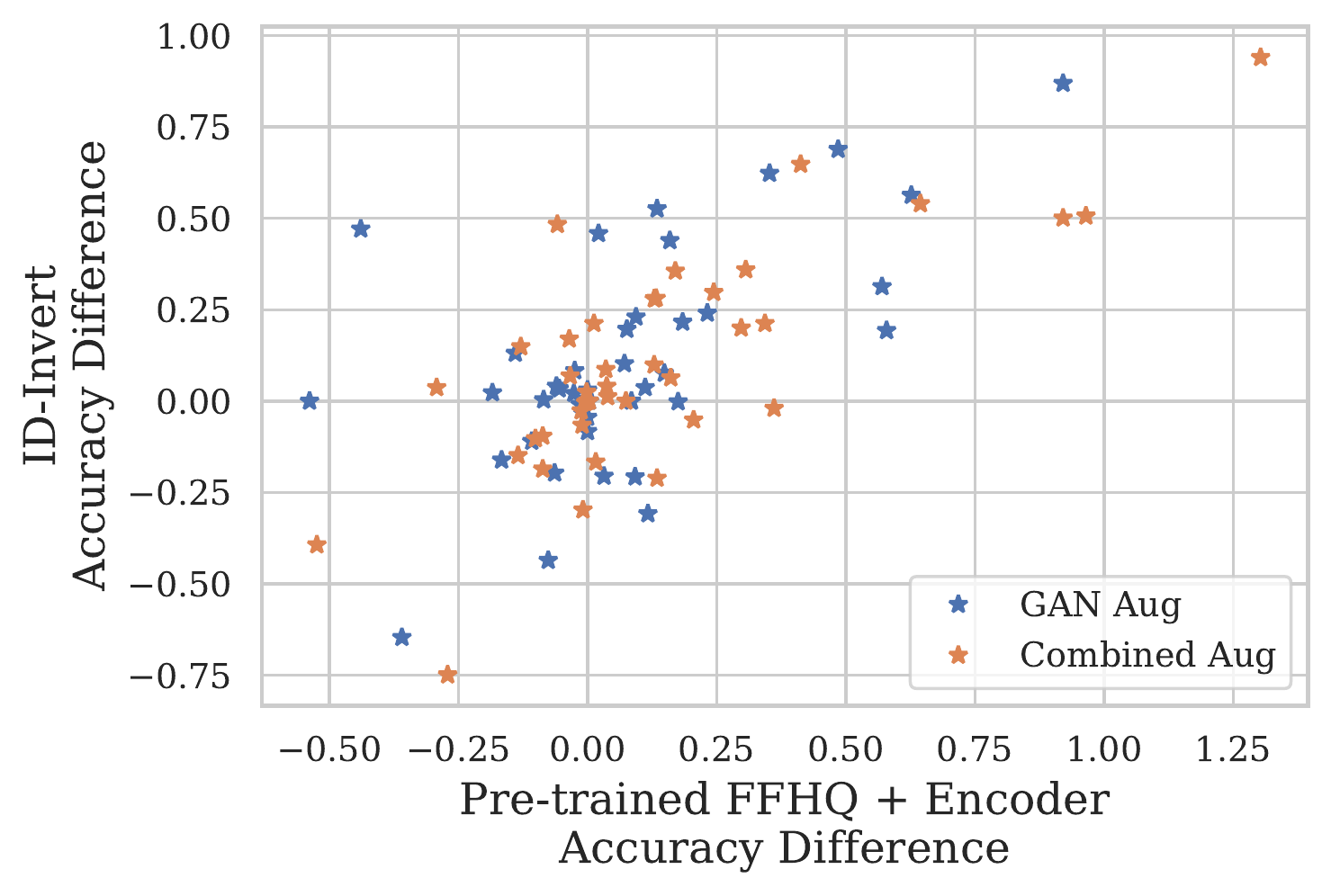}
  \caption{\small Comparison of StyleGAN2~\cite{karras2019analyzing} generator with a separately trained encoder, and the In-domain inversion method~\cite{zhu2020domain}. We plot the accuracy difference between using GAN-based ensembling at test time and standard test accuracy for both methods, where style-mixing augmentation is shown in blue, and combined style-mixing and image augmentations is shown in orange. Over the 40 attribute classification tasks, the accuracy gains of the two methods are similar (Pearson $r=0.69$, $p<0.001$), but the smaller resolution of the In-domain GAN allows for faster inversion and inference.
  \label{fig:sm_idinvert}}
\end{figure}

Rather than using the 1024px pre-trained StyleGAN2~\cite{karras2019analyzing}, we also run the same experiments using a 256px StyleGAN and the In-domain inversion algorithm~\cite{zhu2020domain}. This method is designed for fast image projection, and combined with a smaller resolution GAN, \textit{reduces the computational cost of} inverting real images and creating the GAN-generated variations. Over the 40 facial attributes, the results are correlated (Fig.~\ref{fig:sm_idinvert}). On average over 40 attributes, using the optimization procedure in the main text achieves 0.07 accuracy gain using fine layer style-mix augmentation only at test time, and 0.13 gain using fine style-mix combined with image augmentations at test time; with the In-domain inverter, the average accuracy gain is 0.10 for both fine layer style-mix augmentation only and fine style-mix combined with image augmentations at test time. Thus, we are able to obtain a similar result, but with a lower computational overhead.

\paragraph{Training distributions}

\begin{figure*}[ht!]
  \centering
  \includegraphics[width=0.40\textwidth]{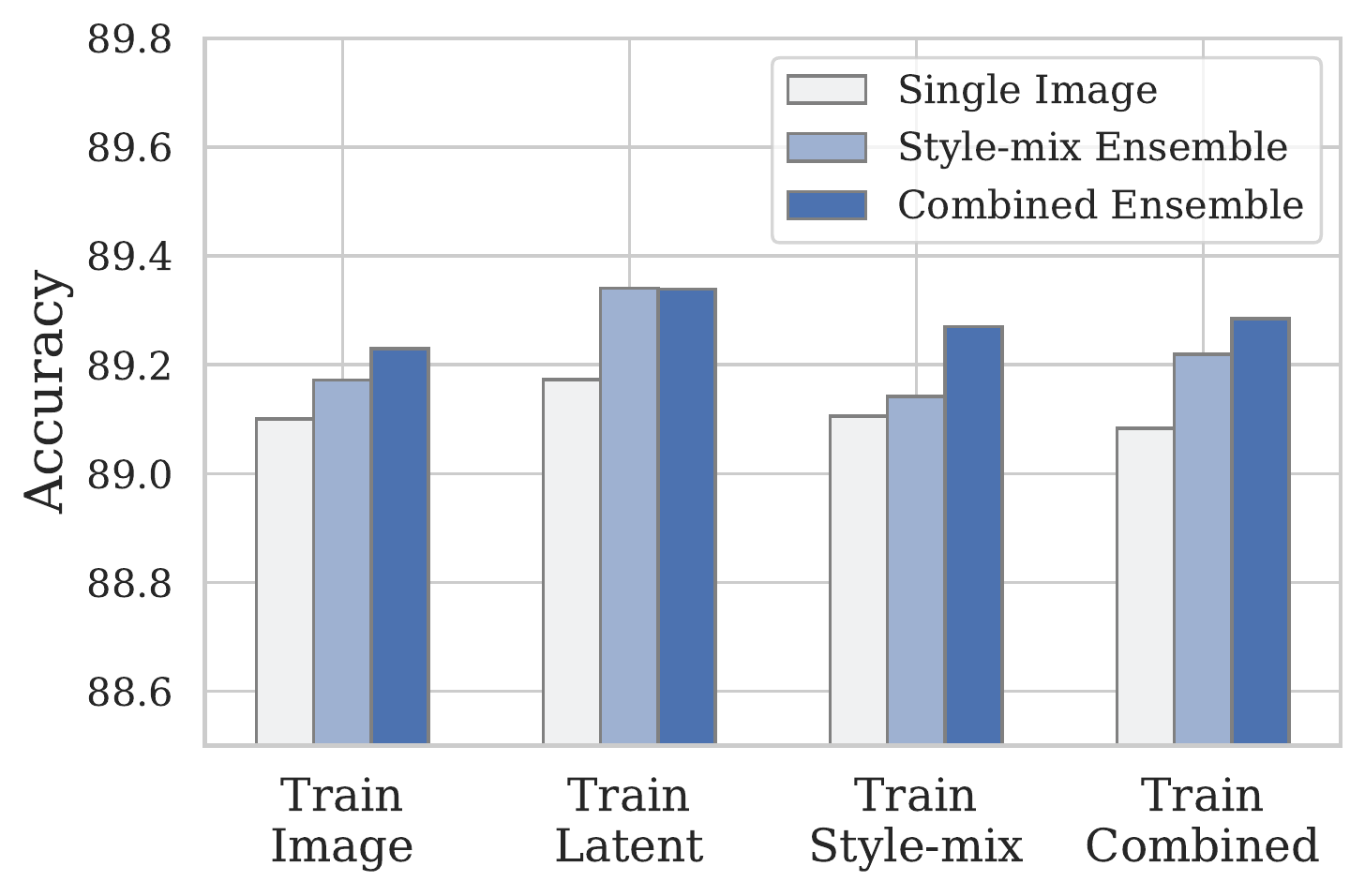}
  \hspace{0.1in}
  \includegraphics[width=0.40\textwidth]{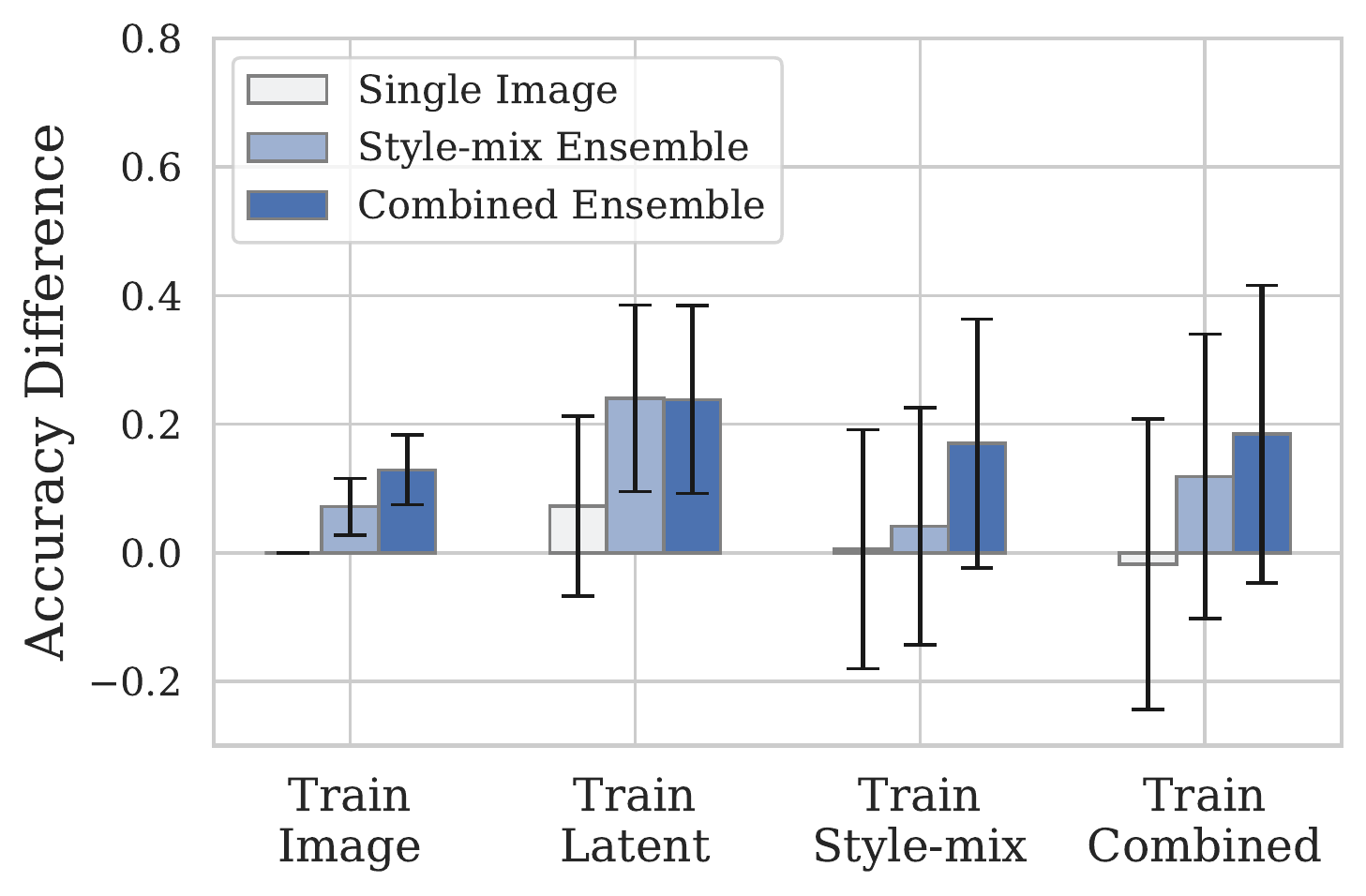}
  \caption{\small Classifier training variations, averaged over 40 facial attributes. (Left) We plot the average classification accuracy when the classifier is trained only on images (Train Image), trained on optimized latent codes corresponding to each image (Train Latent), trained with fine-layer style-mixing (Train Style-mix) and trained with style-mixing and a combination of color and spatial jitter (Train Combined). We evaluate using an ensemble of style-mixed samples, or the combined augmentation (different colored bars). (Right) As there is large variation in the classification accuracy of individual attributes (see Fig.~\ref{fig:sm_faces_distribution}), we also plot the difference in classification accuracy for each setting, compared to a classifier trained on images and evaluated on a single image. While training on the optimized latent codes outperforms training on images, we find that results on style-mixing during training are mixed, as some attributes are sensitive to the style-mixing operation (we show examples in Fig.~\ref{fig:sm_face_train_examples}). Error bars indicate standard error over all 40 attributes.
  \label{fig:sm_face_train_latents}}
  \vspace{-0.1in}
\end{figure*}

\begin{figure*}[ht!]
  \centering
  \begin{subfigure}[t]{0.40\textwidth}
    \centering
    \includegraphics[width=0.9\textwidth]{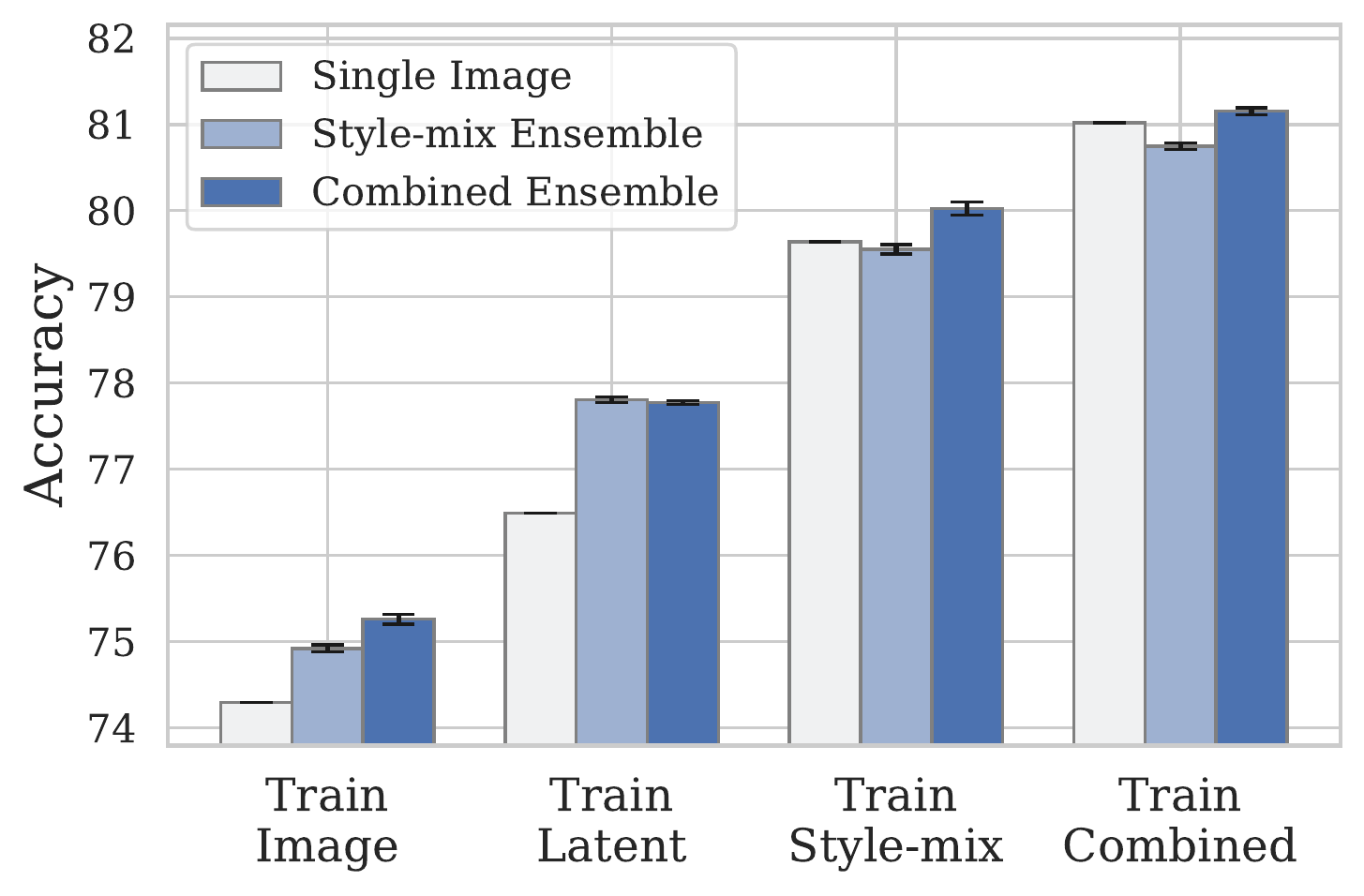}
    \vspace{-3mm}
    \caption{\small Wavy Hair}
  \end{subfigure}
  \begin{subfigure}[t]{0.40\textwidth}
    \centering
    \includegraphics[width=0.9\textwidth]{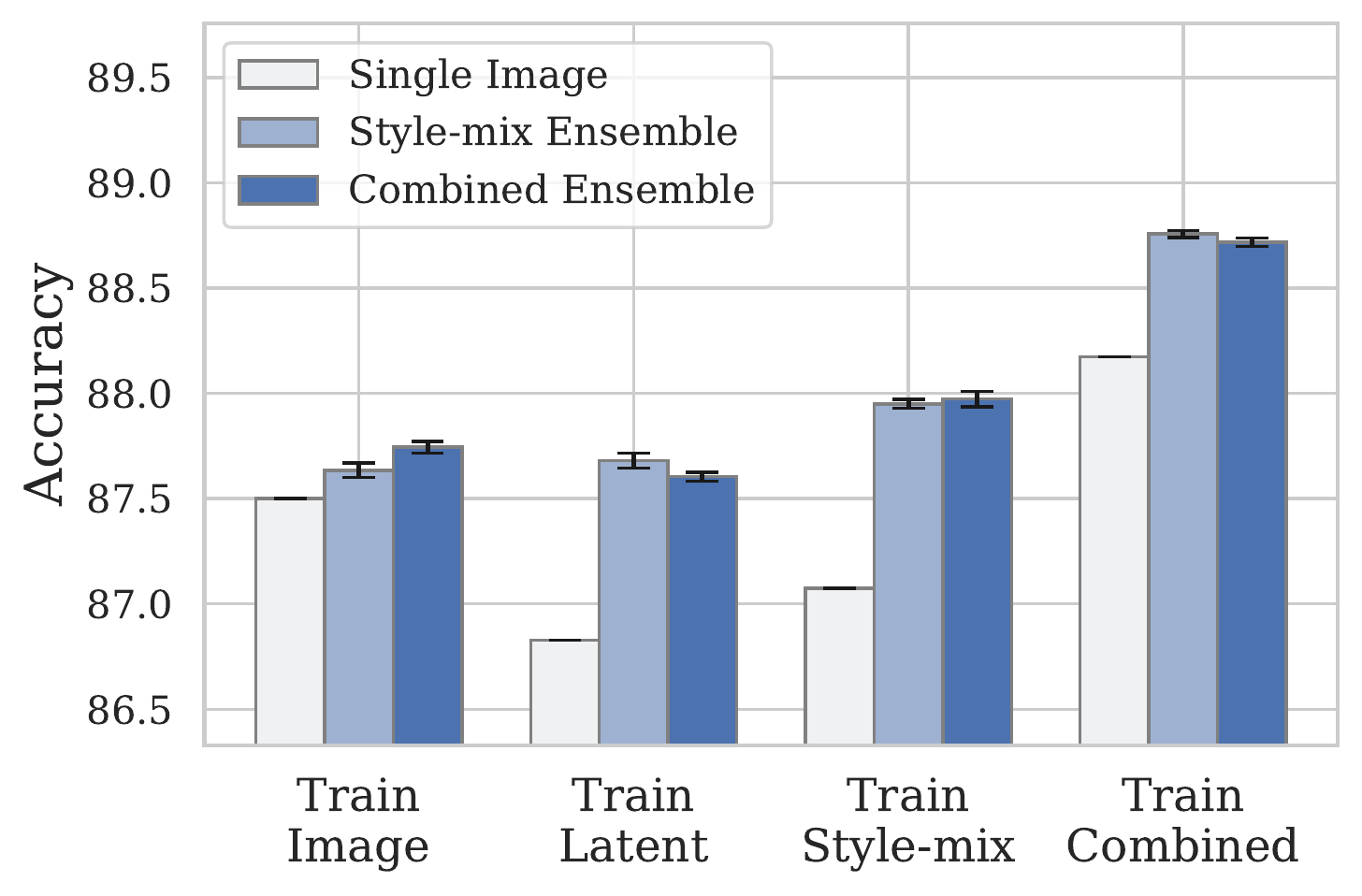}
    \vspace{-3mm}
    \caption{\small Young}
  \end{subfigure}
  \begin{subfigure}[t]{0.40\textwidth}
    \centering
    \includegraphics[width=0.9\textwidth]{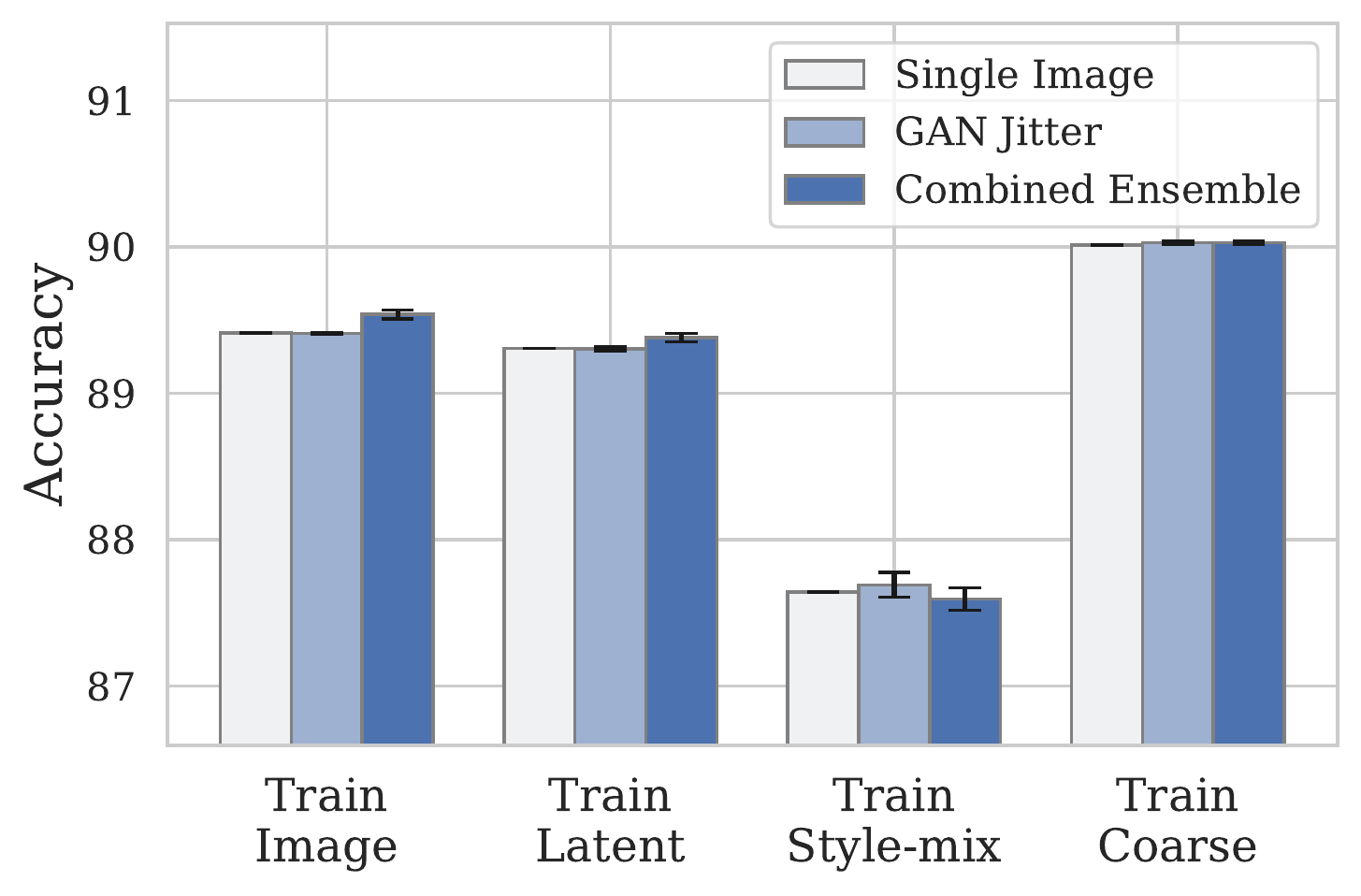}
    \vspace{-3mm}
    \caption{\small Black Hair}
  \end{subfigure}
  \begin{subfigure}[t]{0.40\textwidth}
    \centering
    \includegraphics[width=0.9\textwidth]{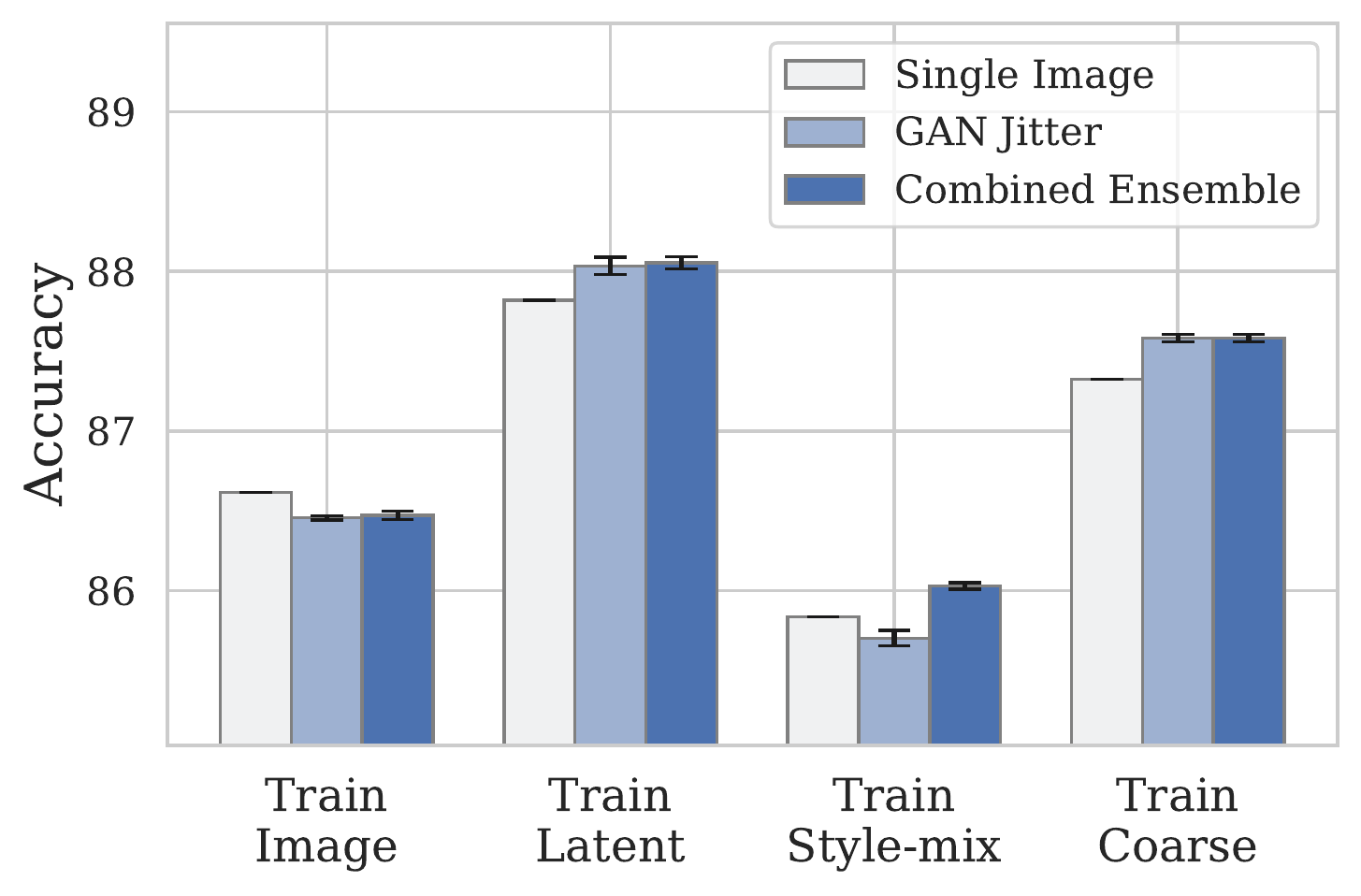}
    \vspace{-3mm}
    \caption{\small Brown Hair}
  \end{subfigure}
  \caption{\small Classifier training variations on individual attributes. (Top) We show two attributes (Wavy Hair and Young) where training with the fine-level style-mixing outperforms training on the original images and the optimized latent code. (Bottom) However, style-mixing during training is harmful for some attributes, such as Black Hair and Brown Hair, where the color adjustment introduced by changing the latent code may create inconsistencies with the image label. For the Black Hair attribute, training with isotropic jittering in coarse layers performs best, while for Brown Hair, training on the latent codes, without additional GAN-based augmentations, is better. For the Black Hair and Brown Hair attributes, we plot test accuracy using coarse-level isotropic jittering as the type of GAN augmentation, which performs better than fine-level style-mixing, except in the case when the classifier is trained with the fine-layer style-mixing augmentation (Train Style-mix). 
  \label{fig:sm_face_train_examples}}
\end{figure*}

Although we find that there are improvements when using GAN-generated views at test time, even when the classifier is only trained on real face images, we also investigate different training variations for the face attribute classifiers (Fig.~\ref{fig:sm_face_train_latents}), such as training the classifier using GAN-reconstructed images and perturbed reconstructions. On average over 40 attributes, we find that adding the style-mixing ensemble is more beneficial when the classifier is also trained on the GAN reconstructions, compared to when these  classifiers are trained only on images. The results of applying fine style-mixing while training these classifiers are mixed and less beneficial than training on the GAN reconstructions alone. This is likely due to the inability of fine-level changes to preserve the the classifier boundaries for certain attributes, such as those based on color. In Fig.~\ref{fig:sm_face_train_examples}, we show examples of two attributes where training with fine-level style-mixing improves classification (Wavy Hair and Young), and two attributes where such an adjustment in the latent code is harmful (Black Hair and Brown Hair). In the latter case, we find that training the classifier with coarse-level isotropic jittering outperforms training the classifier with fine-level style-mixing, as fine-level adjustments in color may create samples inconsistent with the original class label.

\begin{figure*}[ht!]
  \centering
  \includegraphics[width=\textwidth]{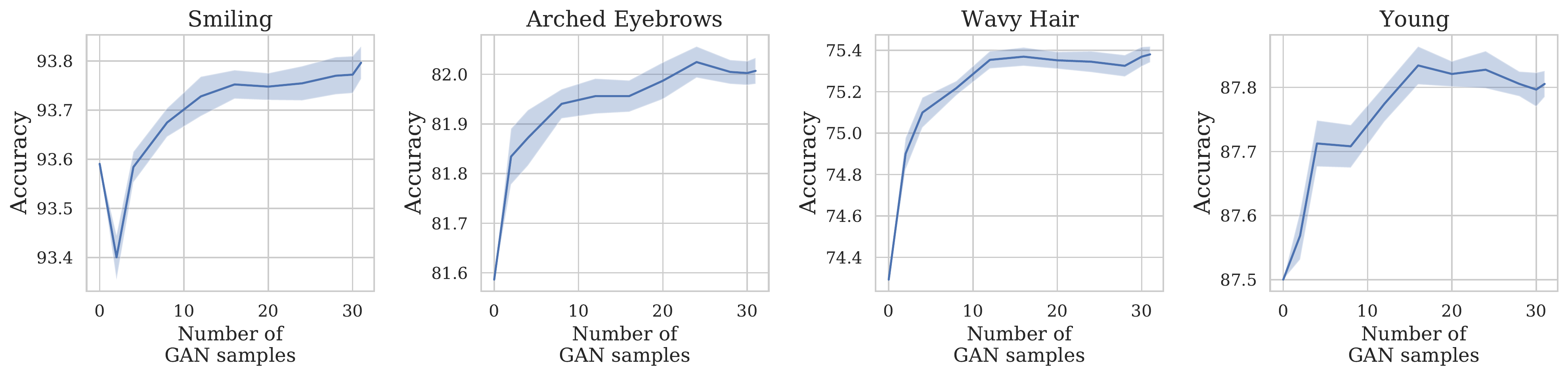}
  \vspace{-2em}
  \caption{\small Effects of ensemble size. Classification accuracy as a function of the number of ensembled deep generative views, on four facial attributes that benefit from GAN-augmented views at test time. Zero views corresponds to using the original input image and adding more views increases accuracy up to a certain point. We use a total of 32 images (1 dataset image and 31 GAN views) in our experiments, as performance saturates. The shaded region corresponds to standard error over random draws of the ensemble elements.
  \label{fig:sm_ensemble_size}}
\end{figure*}

\begin{figure*}[ht!]
  \centering
  \includegraphics[clip, trim=0 0 0 0, width=\textwidth]{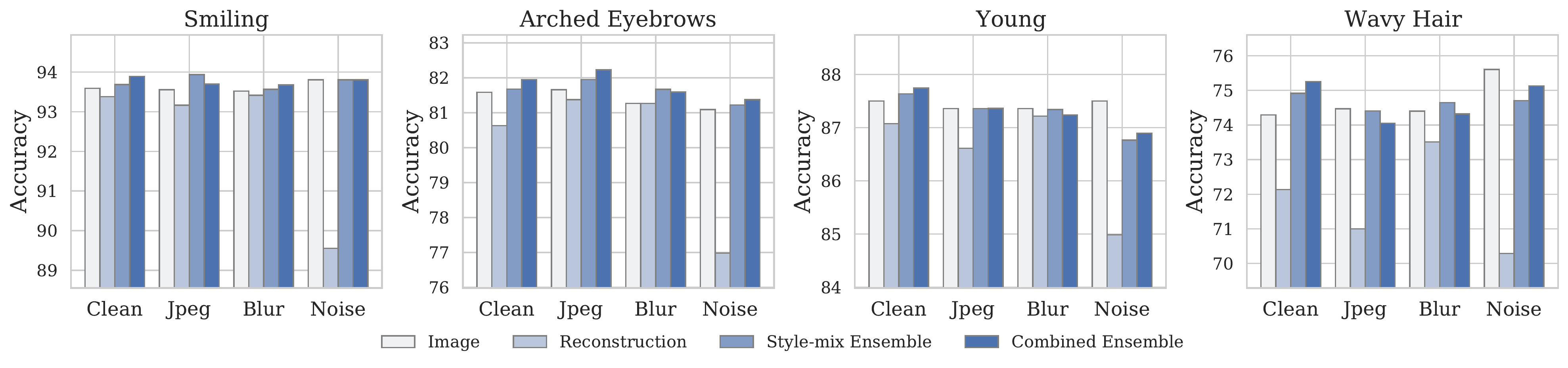}\\
  \includegraphics[width=\textwidth]{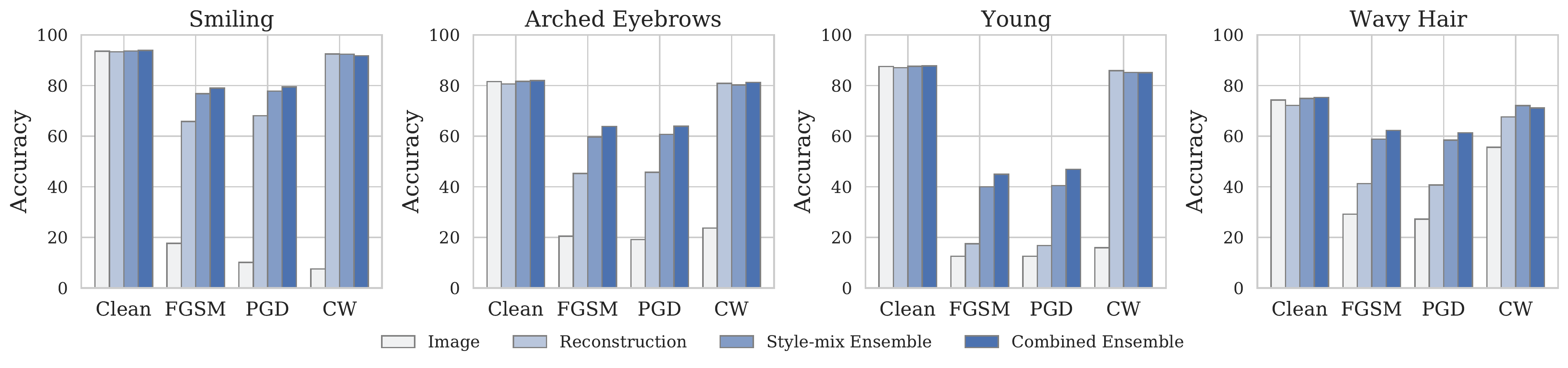}\\
  \vspace{-3mm}
  \caption{\small Robustness to corruptions. We show accuracy on a corrupted image (Image), the GAN reconstruction (Reconstruction), ensembling with GAN style-mixing (Style-mix Ensemble), and ensembling over both traditional and GAN views (Combined Ensemble). (top) On clean images, deep views increase performance across the 4 facial attributes. We test against different types of corruptions: jpeg, Gaussian blur, and Gaussian noise. The results on untargeted corruptions are mixed; in 6 of 12 cases, ensembling improves performance. 
  (bottom) Adversarial attacks (FGSM, PGD, CW) greatly reduce accuracy. On all cases, just GAN reconstruction recovers significant performance. In the majority of cases, ensembling further improves performance.
  \label{fig:sm_corruption}}
\end{figure*}

\begin{figure*}[ht!]
  \centering
  \includegraphics[width=0.9\textwidth]{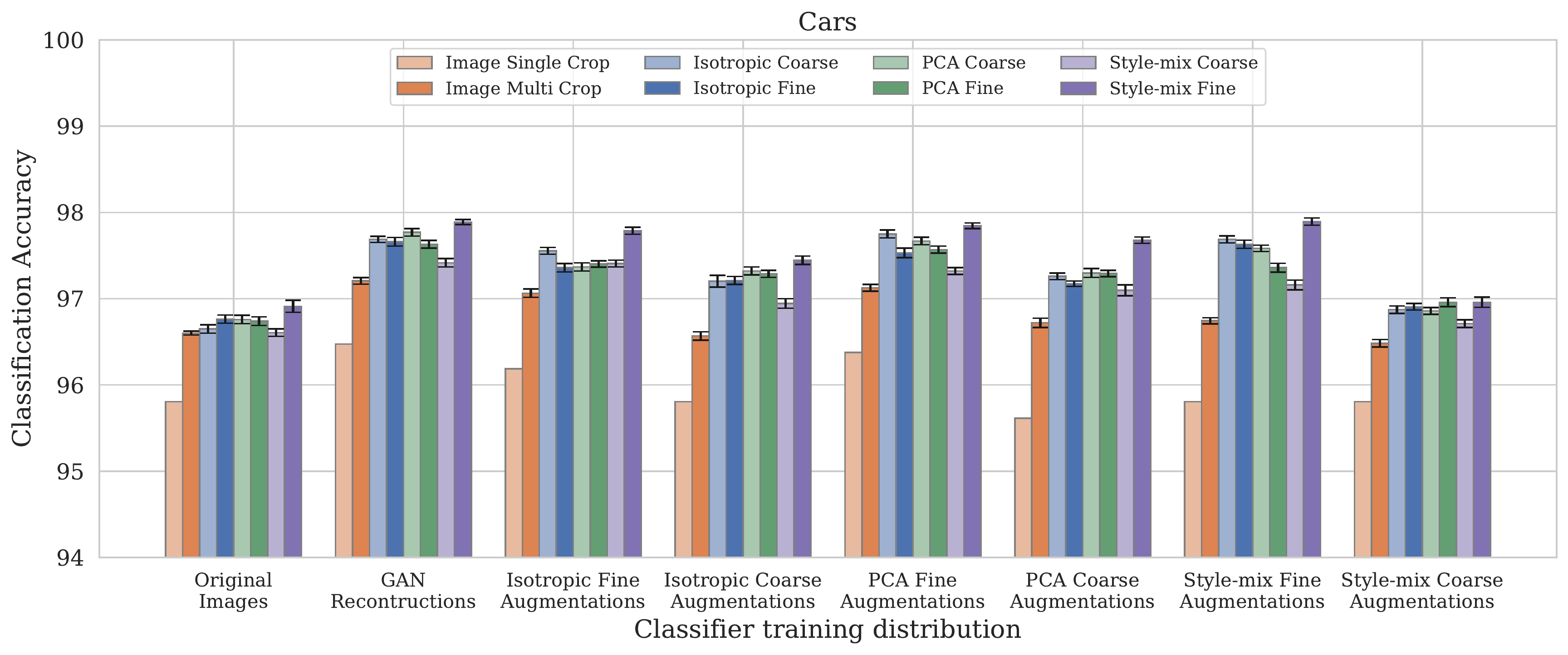}
  \caption{\small Classifier training variations: cars. Starting with a classifier trained on image crops (Original Images), we finetune the classifier using different types of images projected through the generator's latent code -- either GAN Reconstructions or manipulated latent codes using the isotropic, PCA, or style-mixing augmentations at coarse or fine layers during training (along the x-axis). At test time, we then evaluate with an ensemble of different types of GAN perturbations and image crops (different colored bars). On the car domain, we find that adding GAN augmentations only at test time, when the classifier is only trained on dataset images, offers a small increase in accuracy, but there are greater benefits when the classifier is further finetuned on GAN images. In particular, using the fine layer style-mixing augmentation is most beneficial at training and test time. Error bars indicate standard error over 20 bootstrapped samples from 32 ensemble elements.
  \label{fig:sm_car_graphs}}
  \vspace{-0.2in}
\end{figure*}

\paragraph{Additional attributes: ensemble size} When ensembling on face attributes, we use an ensemble size of 32 images. In Fig.~\ref{fig:sm_ensemble_size}, we plot the classification accuracy as a function of the number of ensembled images on four attributes that benefit from test-time ensembling with GAN-generated views (Smiling, Arched Eyebrows, Wavy Hair, and Young). Generally, increasing the number of images in the ensemble improves performance up to a certain point, saturating around an ensemble size of 16 GAN samples.

\paragraph{Additional attributes: image corruptions} We show results on the same four attributes (Smiling, Arched Eyebrows, Wavy Hair, and Young), when the input image is corrupted prior to classification (Fig.~\ref{fig:sm_corruption}). We project the \textit{corrupted} image through the GAN to obtain the reconstructed image, and perform style-mixing on the fine layer to create an ensemble (Style-mix Ensemble). We also ensemble by combining style-mixing and traditional crop and color jittering (Combined Ensemble). On untargeted corruptions, the result of ensembling using GAN augmentations is mixed. Accuracy improves on JPEG and Blur corruptions for the Smiling and Arched Eyebrows attributes, but it does not improve for the Wavy Hair/Young attributes (in fact, Wavy Hair accuracy is higher on the corrupted images than clean images as the classifier is not impaired by the corruption). In the case of Gaussian Noise, classification of the reconstructed image drops, as the projection procedure fails to find a good reconstruction of the noisy image; thus adding GAN perturbations at test time only helps in one out of four attributes (Arched Eyebrows). Note that the classifier is not greatly sensitive to these types of corruptions: classification accuracy between the clean images and the corrupted ones are largely similar, this may be due to the initial downsampling operation on the attribute classifiers following~\cite{karras2019style}, which may reduce the effect of these corruptions. In the targeted corruption setting, the benefits of GAN reconstruction and ensembling is more pronounced. Using the GAN reconstruction rather than the corrupted image increases classification accuracy on these corrupted images, and in the majority of cases, ensembling multiple views from the GAN further improves performance.

\subsection{Additional experiment settings: cars}
\begin{figure*}[ht!]
  \centering
  \begin{subfigure}[t]{0.24\textwidth}
    \centering
    \includegraphics[width=\textwidth]{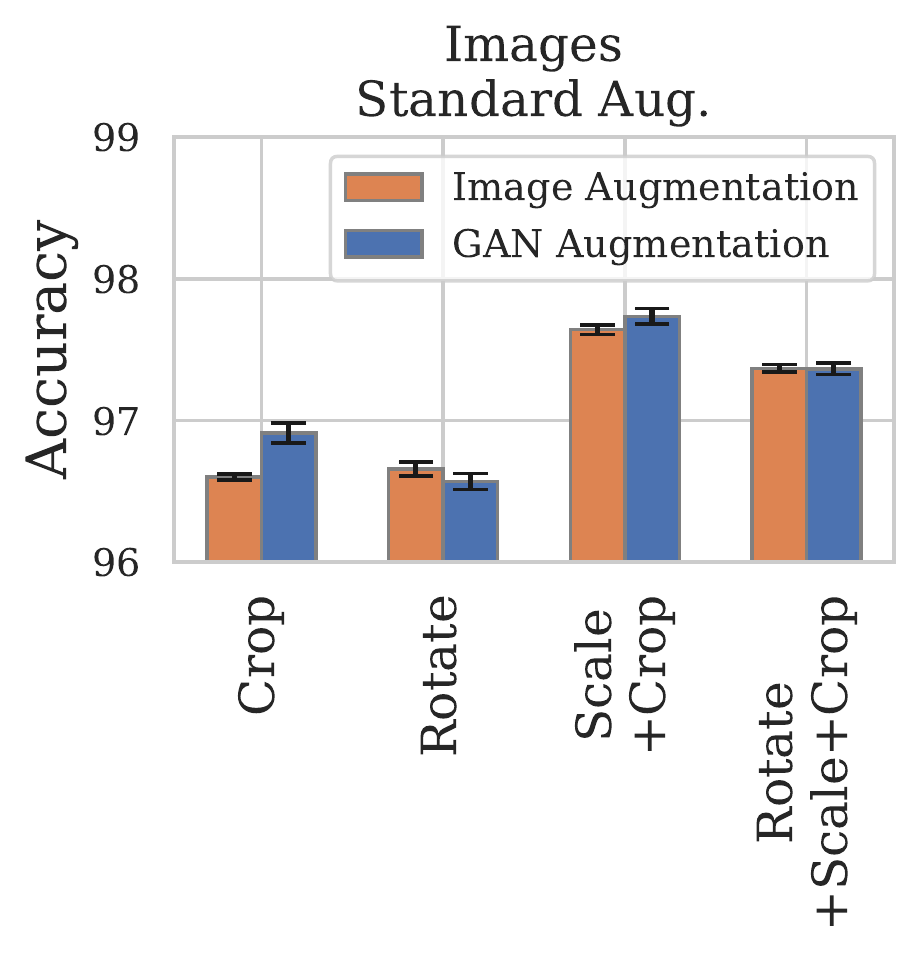}
    \vspace{-5mm}
    \caption{\small }
  \end{subfigure}
  \begin{subfigure}[t]{0.24\textwidth}
    \centering
    \includegraphics[width=\textwidth]{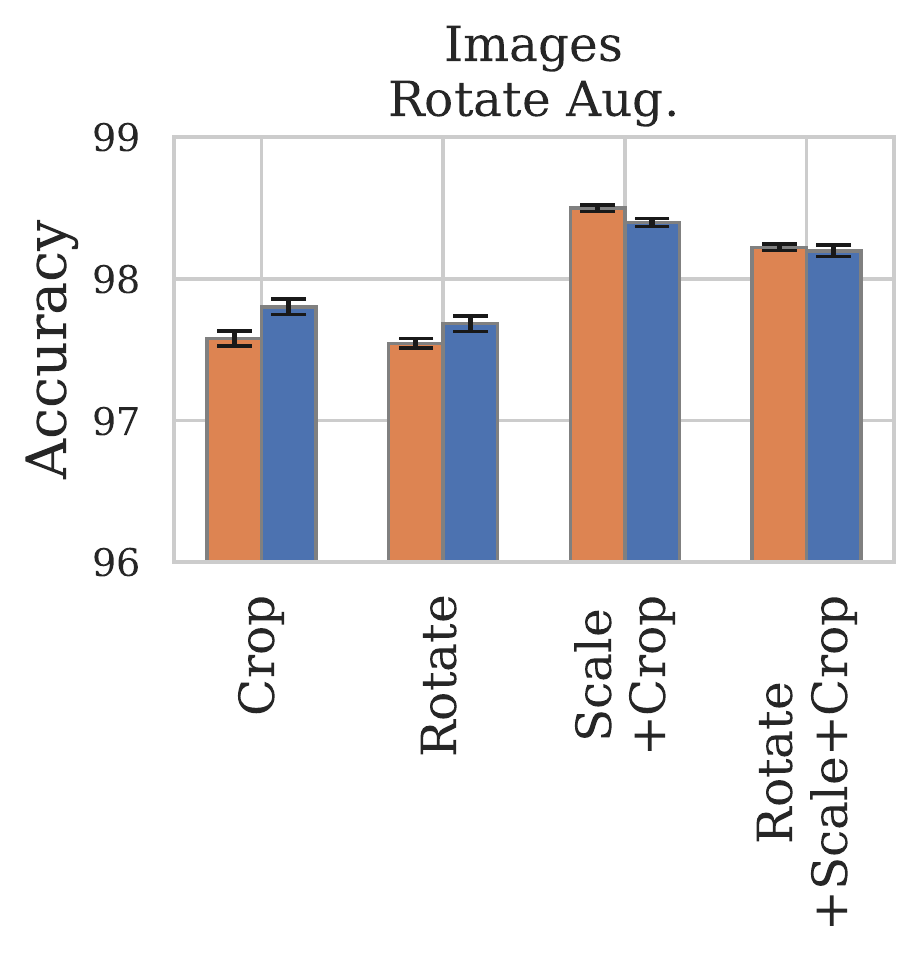}
    \vspace{-5mm}
    \caption{\small }
  \end{subfigure}
  \begin{subfigure}[t]{0.24\textwidth}
    \centering
    \includegraphics[width=\textwidth]{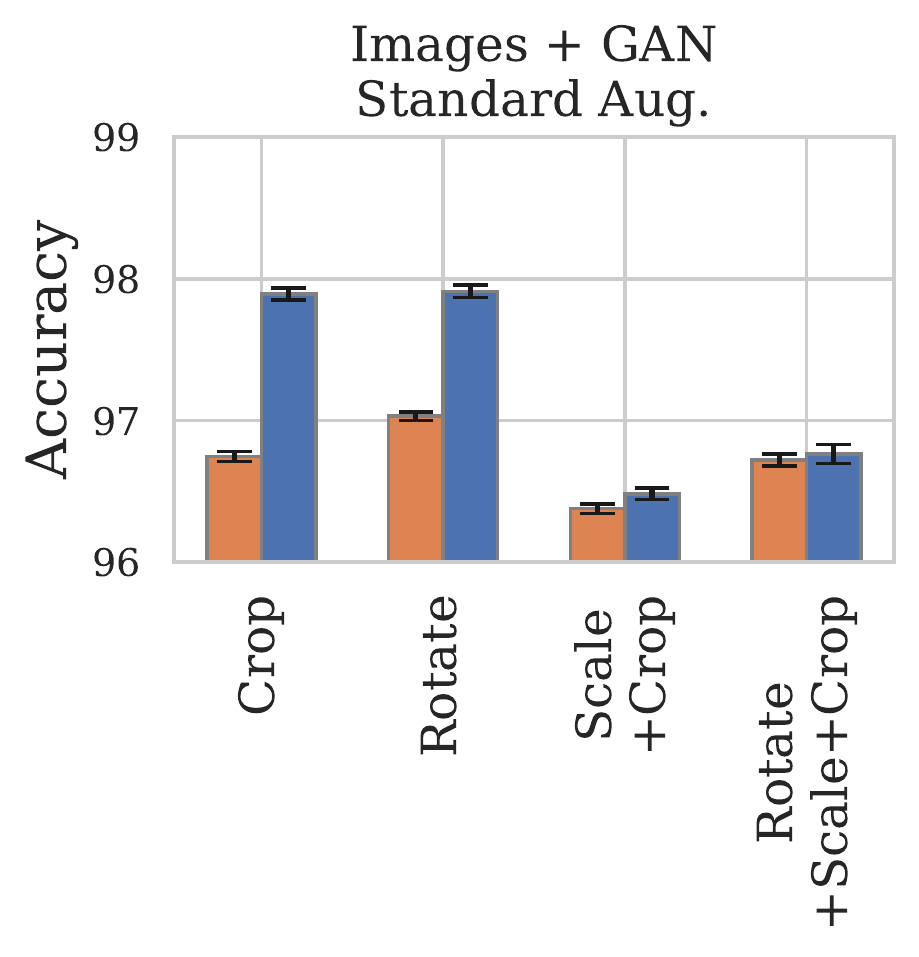}
    \vspace{-5mm}
    \caption{\small }
  \end{subfigure}
  \begin{subfigure}[t]{0.24\textwidth}
    \centering
    \includegraphics[width=\textwidth]{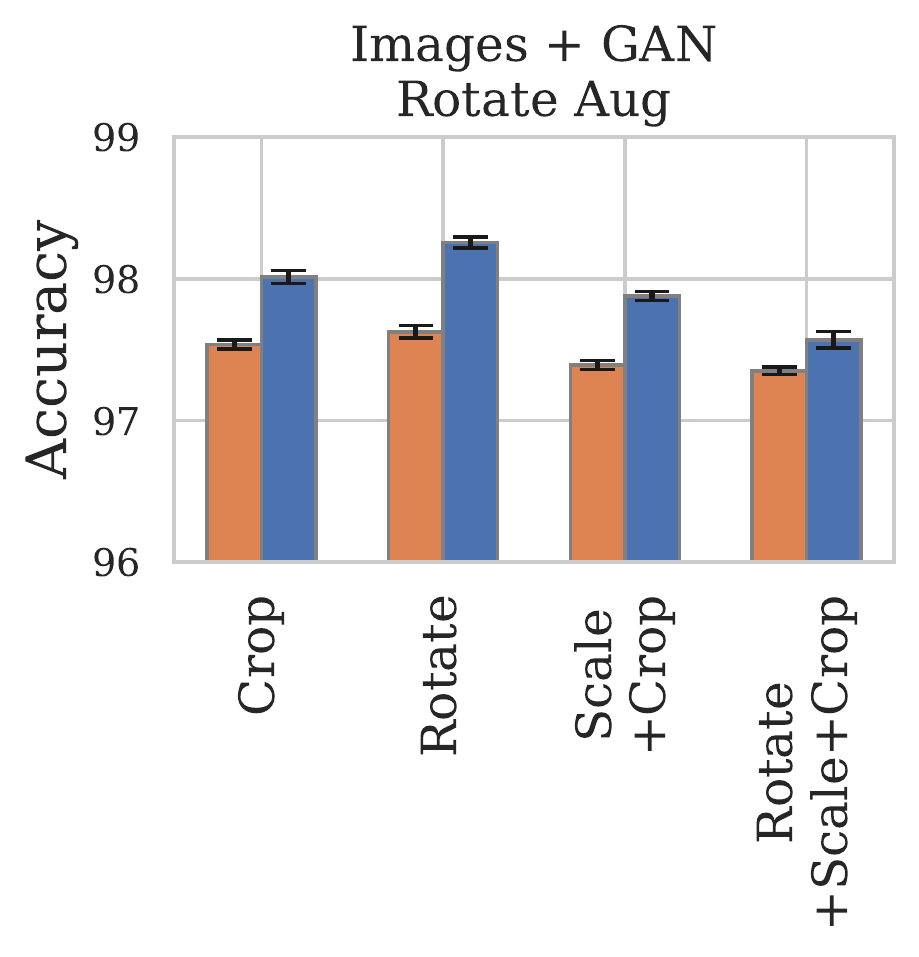}
    \vspace{-5mm}
    \caption{\small }
  \end{subfigure}
  \caption{\small Comparing image and GAN augmentations during training: cars. We investigate the effect of various image augmentations during classifier training and at test time. (a) We first train a classifier on the cars dataset following the ImageNet training transformations, which include random resize, crop, and horizontal flip; applying GAN augmentations at test time can slightly outperform using only image augmentations. (b) Next, we add a random rotate transformation in addition to the previous image transformations during classifier training, and again train on the image dataset; this classifier outperforms the previous classifier, and in this case ensembling images with the scale+crop augmentation at test time is the best. When the classifiers are trained with GAN images, either with the standard training transformations (c) or the additional rotate transformation (d), adding GAN augmentations outperforms using image augmentations at test time, but the classifier's overall accuracy is lower. This suggests that carefully chosen image augmentations during classifier training can still slightly outperform the benefits of GAN-based augmentations at test-time, due to current limitations in image reconstruction using GANs. 
  \label{fig:sm_car_augs}}
\end{figure*}

\begin{figure*}[ht!]
  \centering
  \includegraphics[width=0.9\textwidth]{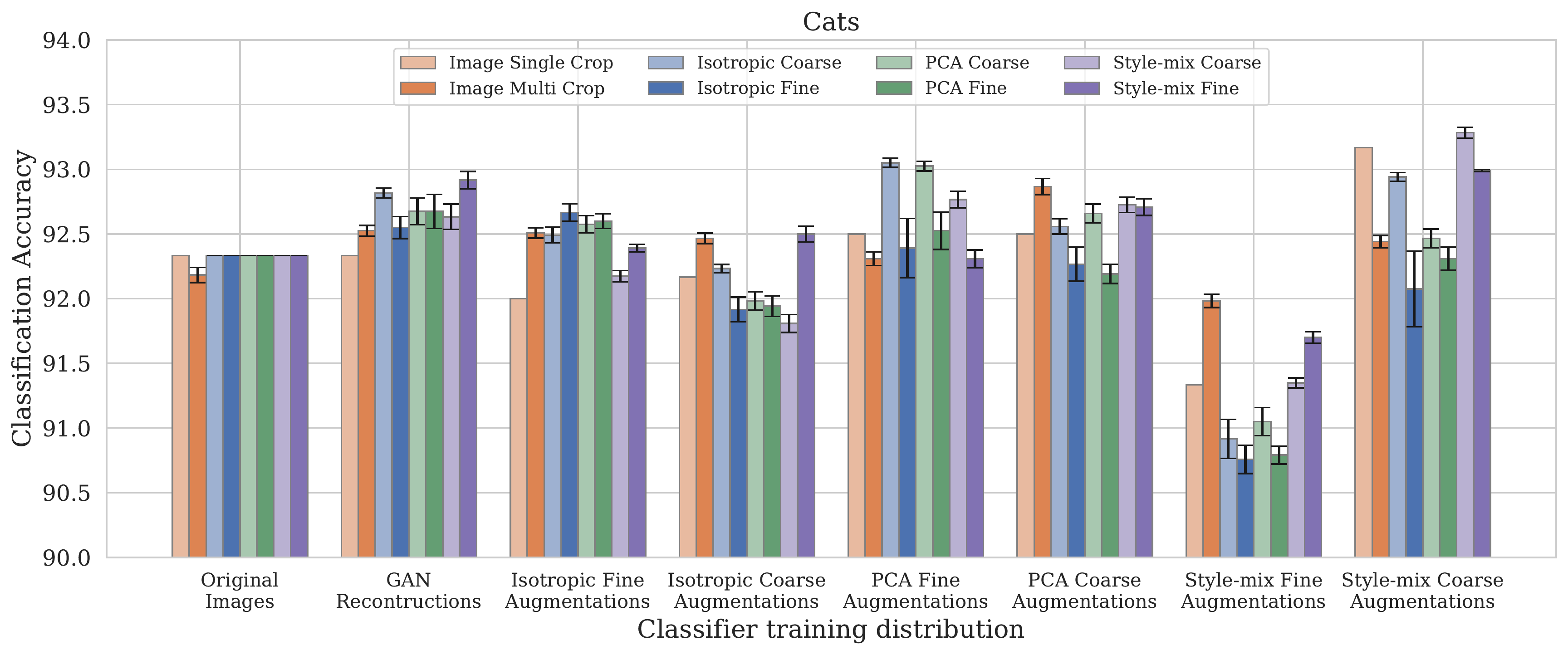}
  \caption{\small Classifier training variations: cats. Similar to the car domain, we start with a classifier trained on image crops (Original Images) and then finetune the classifier using different types of images projected through the generator's latent code -- either GAN Reconstructions or manipulated latent codes using the isotropic, PCA, or style-mixing augmentations at coarse or fine layers during training (along the x-axis). At test time, we then evaluate with an ensemble of different types of GAN perturbations and image crops (different colored bars). On the cat domain, we find that using coarse layer style-mixing offers the highest classification accuracy; it increases image classification accuracy on the test set compared to training the classifier only on image crops, and ensembling using the GAN outputs offers a small additional increase. Error bars indicate standard error over 20 bootstrapped samples from 32 ensemble elements.
  \label{fig:sm_cat_graphs}}
\end{figure*}

\paragraph{Training distributions}

In the main text, we focus on the fine-level style-mixing augmentation when finetuning the classifier on generated samples, which corresponds to small color changes, such as changing the color of the car. In Fig.~\ref{fig:sm_car_graphs} we show results on finetuning the classifier with the remaining isotropic, PCA, and style-mixing augmentations, at both coarse and fine layers. On the car domain, we find that training on fine layer perturbations tend to outperform training on coarse layer perturbations, and at test time, ensembling with fine layer style-mixing augmentations is best. As using multiple image crops in the ensemble consistently increases classification accuracy, we use a combination of 16 image crops and 16 cropped and perturbed GAN reconstructions when ensembling using the GAN output.

\paragraph{Effect of image augmentations} When training the classifier, we use standard random resize, crop, and horizontal flip following the transformations used in ImageNet training\footnote{\url{https://github.com/pytorch/examples/blob/master/imagenet/main.py}}. Here, we also investigate the effect of an additional image rotation augmentation, at both training and test time (we use a random rotation between -10 and 10 degrees prior to the previous transformations of resizing,  cropping, and flipping). When trained on resize, crop, and flip transformations on images, ensembling GAN augmentations outperforms image multi-crop classification using the crop and scale+crop augmentations at test time. Next, we train a classifier using 
rotate, resize, crop, and flip transformations on images; in this case, the using the scale+crop test-time augmentation on images attains the highest accuracy. We then finetune these classifiers by also training on the GAN generated variants: here, combining image and GAN augmentations at test time outperform test-time image augmentations alone, but the accuracy of these finetuned classifiers is lower than the highest accuracy attained by the classifier trained on images with the rotate augmentation (Fig.~\ref{fig:sm_car_augs}). Given the current limitations in GAN reconstruction ability, this suggests that carefully chosen image augmentations \textit{during training} can slightly outperform the benefits of GAN-based augmentations \textit{at test-time}.

\subsection{Additional experiment settings: cats}

\paragraph{Training distributions}

We use a similar setup for the cat classification task as the cars task. Fig~\ref{fig:sm_cat_graphs} shows the classification result, trained on the Original Images (left), and finetuned on GAN reconstructions or each type of perturbation method. In this domain, we find that \textit{training} with the coarse layer style-mix augmentation offers the largest benefit image classification accuracy, and furthermore, ensembling with this same augmentation offers an additional increase in accuracy. Unlike the car domain, the cat images are preprocessed to align and center the face, and so empiricially the benefits of ensembling multiple image crops are less consistent. When ensembling with GAN augmentations, we take the original center-cropped image, and 31 perturbed views from the GAN that are averaged and weighted with ensemble weight hyperparameter $\alpha$.

\begin{figure}[ht!]
  \centering 
  \includegraphics[width=\linewidth]{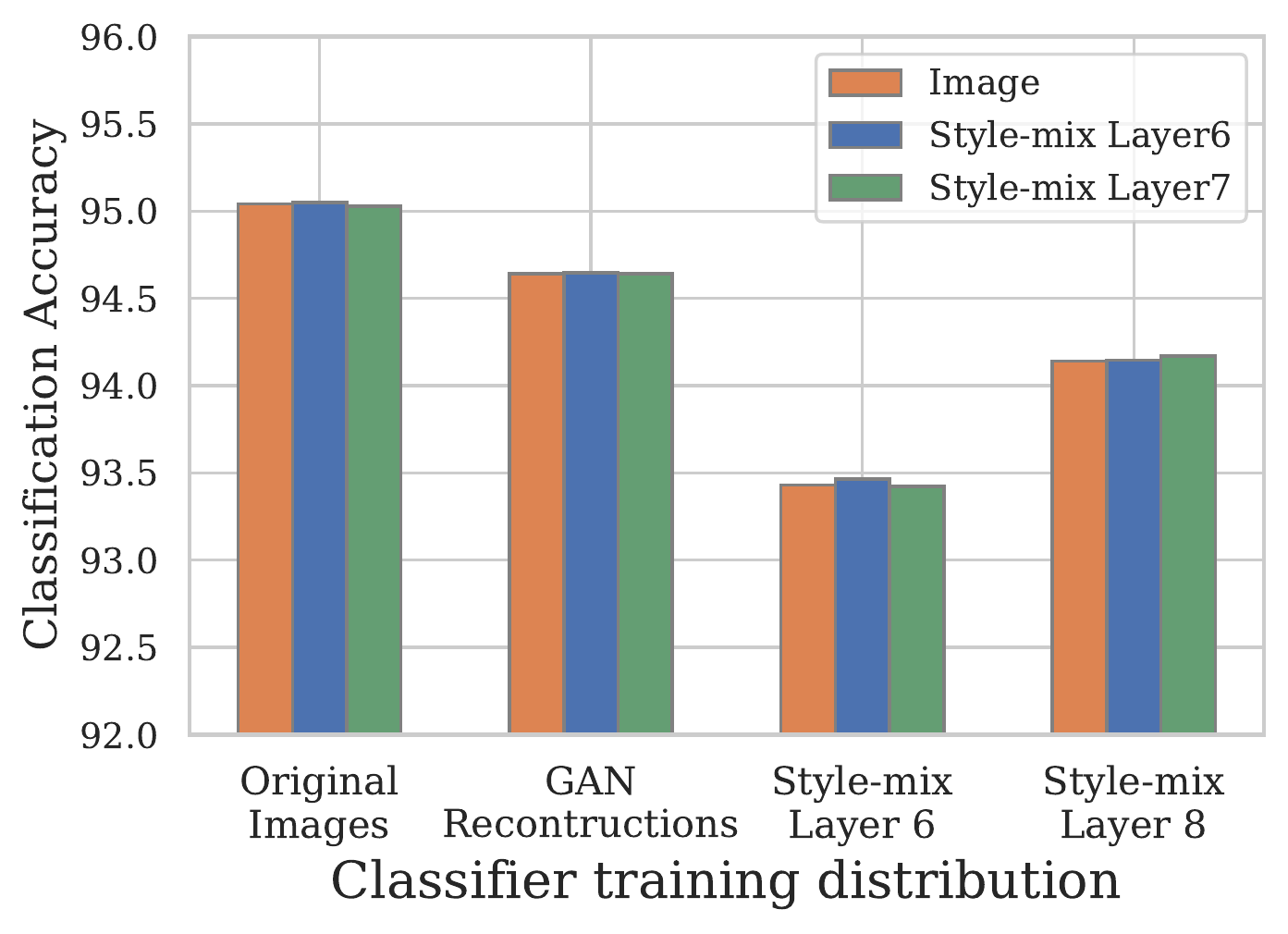}
  \caption{\small Classifier training variations: CIFAR10. With a classifier trained on the CIFAR10 training split, standard test classification accuracy and adding GAN augmentations at test time perform similarly. When the classifier is then finetuned on GAN reconstructions or perturbed GAN reconstructions, adding GAN augmentations at test time offers a small improvement over standard image classification, but the overall classification accuracy is lower, suggesting that the GAN reconstructions cannot perserve the true class well enough to match the classification performanance of the CIFAR10 dataset simages.
  \label{fig:sm_cifar10}}
\end{figure}

\subsection{GAN augmentations with CIFAR10}

\begin{figure*}[ht!]
  \centering
  \begin{subfigure}[t]{0.35\textwidth}
    \centering
    \includegraphics[width=\textwidth]{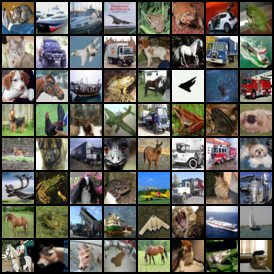}
    \caption{\small CIFAR 10 Images}
  \end{subfigure}
  \begin{subfigure}[t]{0.35\textwidth}
    \centering
    \includegraphics[width=\textwidth]{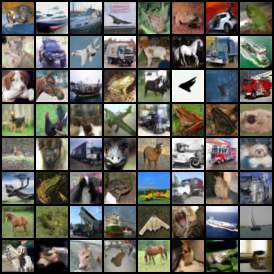}
    \caption{\small GAN Reconstructions}
  \end{subfigure}
  \begin{subfigure}[t]{0.35\textwidth}
    \centering
    \includegraphics[width=\textwidth]{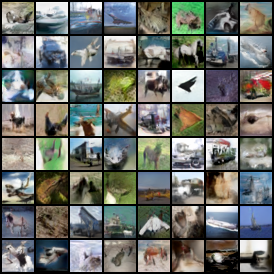}
    \caption{\small Style-mix layer 7}
  \end{subfigure}
  \begin{subfigure}[t]{0.35\textwidth}
    \centering
    \includegraphics[width=\textwidth]{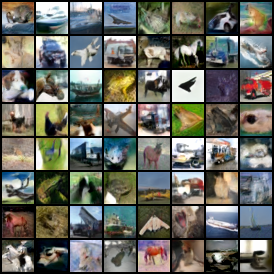}
    \caption{\small Style-mix layer 8}
  \end{subfigure}
  \caption{\small Qualitative examples of CIFAR10 GAN reconstructions. (a) CIFAR10 images from the test set, (b) the GAN reconstructions of the test images (c) swapping the reconstructed latent code with a random latent code from the same predicted class (style-mixing) at layer 7, and (d) style-mixing at the final layer.
  \label{fig:cifar10_qualitative}}
  \vspace{-0.2in}
\end{figure*}

We use a class-conditional StyleGAN2~\cite{karras2020training} to conduct preliminary experiments on the CIFAR10 dataset~\cite{krizhevsky2009learning} (we also tried the unconditional CIFAR StyleGAN, but obtained poorer reconstructions). We first reserve the final 5000 training images for validation, and train a Resnet-18 classifier on the remaining 45000 training images, which achieves accuracy of 95.04\% on the CIFAR10 test set\footnote{\url{https://github.com/kuangliu/pytorch-cifar}}. To project the 32x32 images into the GAN latent space, we first predict the class of each image, use the average $W$ latent of the predicted class to initialize optimization, and optimize for 200 steps, taking about eight seconds per image. We show qualitative examples of CIFAR10 test images, and their GAN reconstructions, and the result of swapping a random latent code, i.e. style-mixing, corresponding to the same predicted class at the seventh and eighth layers in Fig.~\ref{fig:cifar10_qualitative} (the GAN has a total of eight layers). When the classifier is trained only on real images, we find that classifying the reconstructed test images is harder than the real images (accuracy drops from 95.04\%  to 84.68\%). With the ensemble weighting hyperparameter $\alpha$, we find that while the validation split has a small increase in classification accuracy of 0.04\%, although when applying this same ensemble weight to the test split, the accuracy increase is only 0.01\%. However, using the \textit{optimal} $\alpha$ weight on the test split increases accuracy by 0.05\%. We then finetune the classifier on GAN reconstructions of the training set, and additionally perform style-mixing in the seventh and eighth layers. When trained with style-mixing, adding GAN-generated views at test time can outperform standard image classification, but note that the overall accuracy of the classifier is lower (Fig.~\ref{fig:sm_cifar10}). These initial results suggest that the GAN reconstructions currently not perserve the true class well enough to attain the same performance as classification of CIFAR10 images, and due to the smaller resolution of the CIFAR10 StyleGAN, the style-mixing operation in later layers may not be sufficiently disentangled from class identity to offer benefits when ensembling at test time, comparing to classifying images directly.

\end{document}